\documentclass[ijoc,nonblindrev]{informs3} 

\OneAndAHalfSpacedXI 

\usepackage{endnotes}
\let\footnote=\endnote
\let\enotesize=\normalsize
\def\notesname{Endnotes}%
\def\makeenmark{$^{\theenmark}$}
\def\enoteformat{\rightskip0pt\leftskip0pt\parindent=1.75em
  \leavevmode\llap{\theenmark.\enskip}}

\usepackage[dvipsnames]{xcolor}
\usepackage{mwe}
\usepackage{tikz}
\usepackage[abs]{overpic}
\usetikzlibrary{arrows.meta,decorations.pathreplacing,angles,quotes}
\usepackage{longtable}
\usepackage{enumerate}
\usepackage{mathrsfs,amsfonts}
\usepackage{lscape}
\usepackage{subcaption}
\usepackage[absolute,overlay]{textpos}
\usepackage{dsfont}
\usepackage[margin=1in]{geometry}
\usepackage{booktabs}
\usepackage{comment}
\usepackage{transparent}
\usepackage[normalem]{ulem}
\usepackage{mathtools}
\usepackage{multirow}
\usepackage{multicol}
\usepackage{float}
\usepackage{arydshln}
\usepackage{pgfplots}
\pgfplotsset{compat=newest,ticks=none}
\usepgfplotslibrary{patchplots}
\usepackage{makecell}
\usepackage{enumitem}
\usepackage{multicol}
\usepackage{romannum}

\newlist{deflist}{description}{2}
\setlist[deflist]{labelwidth=2cm,leftmargin=!,font=\normalfont}

\newcommand{\flowoct}{\operatorname{FlowOCT}}
\newcommand{\bendersoct}{\operatorname{BendOCT}}
\newcommand{\mcftwo}{\operatorname{MCF2}}
\newcommand{\mcfone}{\operatorname{MCF1}}

\newcommand{\cutone}{\operatorname{CUT1}}
\newcommand{\cuttwo}{\operatorname{CUT2}}
\newcommand{\base}{\operatorname{BASE}}
\newcommand{\baseq}{\operatorname{BASE+q}}
\newcommand{\baseP}{\mathscr{P}_{\operatorname{BASE}}}
\newcommand{\baseqP}{\mathscr{P}_{\operatorname{BASE+q}}}
\newcommand{\flowoctP}{\mathscr{P}_{\operatorname{FlowOCT}}}
\newcommand{\mcfoneP}{\mathscr{P}_{\operatorname{MCF1}}}
\newcommand{\mcftwoP}{\mathscr{P}_{\operatorname{MCF2}}}
\newcommand{\cutoneP}{\mathscr{P}_{\operatorname{CUT1}}}

\newcommand{\cuttwoP}{\mathscr{P}_{\operatorname{CUT2}}}
\newcommand{\child}{\operatorname{CHILD}}

\newcommand{\milo}{\operatorname{MILO}}

\newcommand{\dist}{\operatorname{dist}}
\newcommand{\proj}{\operatorname{proj}}

\usepackage{natbib}
 \bibpunct[, ]{(}{)}{,}{a}{}{,}%
 \def\bibfont{\small}%
\TheoremsNumberedThrough     
\ECRepeatTheorems

\EquationsNumberedThrough    

\begin{document}


\RUNAUTHOR{Alston, Validi, \& Hicks}

\RUNTITLE{MILO formulations for binary classification trees}

\TITLE{Mixed Integer Linear Optimization Formulations for \\ Learning Optimal Binary Classification Trees}
\ARTICLEAUTHORS{
\AUTHOR{Brandon Alston, Illya V. Hicks}
\AFF{Computational Applied Mathematics and Operations Research, Rice University, Houston, TX 77005,
\EMAIL{\{bca3, ivhicks\}@rice.edu}}
\AUTHOR{Hamidreza Validi}
\AFF{Industrial, Manufacturing \& Systems Engineering, Texas Tech University, Lubbock, TX 79409,
\EMAIL{hvalidi@ttu.edu}}
}

\ABSTRACT{Decision trees are powerful tools for classification and regression that attract many researchers and industry professionals working in the burgeoning area of machine learning. In particular, decision trees provide interpretability, which is often preferred over other higher-accuracy methods that are relatively uninterpretable. An optimal binary classification tree has two types of vertices, (i) branching vertices which have exactly two children and where datapoints are assessed on a set of discrete features and (ii) leaf vertices at which datapoints are given a discrete prediction, and can be obtained by solving a biobjective optimization problem that seeks to (i) maximize the number of correctly classified datapoints and (ii) minimize the number of branching vertices. In this paper, we propose four mixed integer linear optimization (MILO) formulations for designing optimal binary classification trees: two flow-based formulations and two cut-based formulations. We show theoretical improvements on the strongest flow-based MILO formulation currently in the literature and conduct experiments on 16 publicly available datasets to show our models' ability to scale, strength against traditional branch and bound approaches, and robustness in out-of-sample test performance. Our code and data are available on GitHub.
}%



\maketitle

%

\section{Introduction}\label{introduction}
Decision trees have been employed as tools in various applications including decision making in management science~\citep{magee1964} and solving integer optimization problems in operations research~\citep{land1960} since the 1960s. With the emergence of machine learning around 1980,~\citet{breiman1984} applied decision trees to classification and regression problems.
Due to their interpretability, binary decision trees are employed in a wide range of applications, including but not limited to healthcare~\citep{yoo2020, li2021, albaqami2021}, geological surveying~\citep{balk2000}, cyber-security~\citep{maturana2011, kumar2013}, financial analysis~\citep{delen2013, charlot2014, manogna2021}, and more recently fair decision making~\citep{ntoutsi2019, barata2021, valdivia2021}. Nearly all such applications employ decision trees for feature selection of training data and classification/regression of test data resulting in a supervised machine learning method. Their interpretability is due to their lack of a \emph{black box} nature and easy to understand branching rules and structure. This interpretability is often preferred over higher-accuracy black-box methods. Optimal binary classification decision trees select binary tests to perform at each branching vertex and classes to assign to each leaf vertex to maximize prediction accuracy or minimize misclassification rates.~\citet{hyafil1976} show building an optimal decision tree is NP-hard and heuristic algorithms were first proposed to find approximations of decision trees as computer technology was not advanced enough to efficiently solve exact algorithms in the 1980s. 
Comparisons of heuristic tree models provide useful outlines for selecting attributes given specific conditions, but~\citet{murthy2004} concludes no one heuristic for splitting is superior.
Recently, optimization solvers such as Gurobi and CPLEX, have become substantially more powerful (speedup factor of 450 billion over a 20 year period) through advancements in hardware, effective use of cutting plane theory, disjunctive programming for branching rules, and improved heuristic methods, as detailed by~\citet{bixby2012}.
These advancements simultaneously eliminate the practical barrier for implementing Mixed Integer Linear Optimization ($\milo$) formulations to solve NP-hard problems and the prejudice relevant during the inception of exact algorithms for the optimal decision tree problem, as noted by~\citet{bertsimas2017}. 

\textbf{Our Contribution}: In this paper, we focus on univariate binary classification decision trees: trees in which each parent has exactly two children and each branching vertex observes a single feature. We propose four $\milo$ formulations for finding optimal binary decision trees and show their strong linear optimization (LO) relaxations compared to current $\milo$ formulations in the literature. Further we show total unimodularity properties of our proposed models. Through experimental testing on 16 publicly available datasets, we highlight the practical application of the proposed $\milo$ formulations, their ability to scale, and strong performance against traditional branch and bound methods. In Section~\ref{related-work}, we review related work on binary decision trees. In Section~\ref{milosection}, we propose our four $\milo$ formulations: two multi-commodity flow formulations ($\mcfone$ and $\mcftwo$), and two cut-based formulations ($\cutone$ and $\cuttwo$) for finding optimal binary trees. In Section~\ref{theorycomp}, we provide theoretical comparisons against the most recent and strongest $\milo$ formulation, $\flowoct$, proposed by~\citet{aghaei2021}. In Section~\ref{tu} we show total unimodularity properties through equitable bicolorings of column submatrices with the results of~\citet{ghoulia1962}. In Section~\ref{compexpr}, we provide computational experiments supporting our theoretical results and report in-sample optimization performance, out-of-sample test performance, efficiently generated Pareto frontiers for an understanding of the relationship between tree topology and out-of-sample test performance, and variations on our proposed cut-based $\milo$ models for speeding up solution time. Our goal is to provide those interested in finding optimal binary decision trees a set of implementable and flexible $\milo$ formulations.

\section{Related Work on Binary Decision Trees}\label{related-work}

Various mathematical optimization techniques have been applied to solve the binary decision tree problem.~\citet{murthy1994} employ hill-climbing techniques paired with randomization.~\citet{orsenigo2003} use discrete SVM operators counting misclassified points rather than measuring distance at each node of the tree; sequential LP-based heuristics are then employed to find the complete tree.~\citet{menze2011} extend oblique random forests with linear discriminate analysis (LDA) to find optimal internal splits.~\citet{wang2015} apply logistic regression to find branching hyperplanes while maintaining sparsity through a weight vector.~\citet{wickramarachchi2016} employ Householder transformations. More recently augmented machine learning techniques have been employed to build decision trees.~\citet{kontschieder2015} combine convolutional neural nets and stochastic, differentiable decision trees to find global tree parameters.~\citet{ balestriero2017} uses a modified hashing neural net framework with sigmoid activation functions and independent multilayer percepetrons that are equivalent to vertices of a decision tree.~\citet{yang2018} use a one-layer neural network with a softmax activation function and a \emph{soft} binning function for branching.~\citet{zantedeschi2020} employ stochastic descent for branching attributes, auxiliary variables for linearity, and a unique tree-structured isotonic optimization algorithm for pruning-aware decision trees.~\citet{lee2020} show decision trees are related to piece-wise linear neural nets with locally constant gradients.

Researchers also apply customized dynamic programming, often branch-and-bound processes, to combat searching over large spaces associated with finding optimal decision trees.~\citet{nijssen2007} optimize a ranking function under topology constraints.~\citet{aglin2020} use two branch-and-bound approaches that cache itemsets used for cutting the search space and only including vertices not in the cache in the branch-and-bound cuts.~\citet{angelino2017} create optimal decision lists which applied to tree structures through bit-vector libraries and specialized data structures.~\citet{demirovic2022} introduce constraints on the depth and number of nodes to combat scaling issues.~\citet{mctavish2022} employ guessing strategies related to feature binarization, tree depth, and bound tightening while optimizing misclassification loss and a sparsity penalty over leaves.~\citet{mazumder2022} explore the search space through the quantiles of the features', which can be continuous, distributions. We would like to note formulations such as~\citet{michini2020,nijssen2007,bertsimas2017,breiman1984,demirovic2022,mazumder2022} allow for multivariate branching; branching vertices act as separating hyperplanes. Multivariate branching rules provide more flexibility than univariate branching rules, resulting in sparser trees. Objective functions of multivariate branching rules often involve weighting between hyperplane complexity and classification metrics. However such formulations are out of the scope of this paper due to the plethora of methods involved for finding hyperplanes which include but are not limited to SVM's~\citep{boser1992}, Lagrangian operators, and Lenstra's algorithm~\citep{lenstra1987}.

\citet{breiman1984} note continued growth of the tree is indicative of successful splits and the growth itself is a one-step optimization problem; thus, using unnatural objective functions related to branching rather than classification metrics was acceptable.~\citet{bertsimas2017} emphasize steps of building a decision tree involve discrete decisions (which vertex to split on? which variable to split with?) and discrete outcomes (is a datapoint correctly classified? which leaf does a datapoint end on?). Therefore, one should consider building optimal decision trees using $\milo$ formulations. To the best of our knowledge,~\citet{bennett1996} propose the first $\milo$ formulation for designing optimal multivariate (branching vertices observe more than one feature) decision trees. They fix the structure of the tree, the number of branching vertices and the classes of leaf vertices before solving. They showed that optimal solutions of their problem lie on the extreme points of the feasible set, even with a nonconvex objective function. They also established branching decisions as sets of disjunctive linear constraints~\citep{bennett1995, mangasarian1993}.~\citet{bertsimas2007} introduced CRIO as a software package that employs CPLEX 8.0 for solving optimization models.~\citet{bertsimas2017} propose OCT which outperforms CART~\citep{breiman1984} in accuracy.~\citet{narodytska2018} present a scalable SAT-based approach to build depth $D$ trees. They generate valid binary encodings for a specified number of vertices $N \leq D$, such that valid binary trees of size $N$ can be generated; they also provide upper bounds on $N$.~\citet{verwer2019} propose BinOCT, a binary-linear programming model aiming to reduce the dependence of the problem size on the size of the training dataset.~\citet{gunluk2018} and~\citet{firat2020} both propose column generation approaches.~\citet{gunluk2021} formulate IP models for decision trees with categorical data. Optimal randomized classification trees (ORCT) from~\citet{blanquero2021} uses a continuous optimization method for learning trees by replacing discrete binary decisions in traditional trees with probabilistic decisions.

Recently,~\citet{aghaei2021} propose a $\milo$ formulation whose LP relaxation is at least as strong as that of OCT~\citep{bertsimas2017} and BinOCT~\citep{verwer2019}. They modify the structure of a traditional decision tree by adding a single source vertex and a single sink vertex to turn the tree into a \emph{directed acyclic graph}.~\citet{aghaei2021} also propose two novel indices measuring disparate impact and disparate treatment for learning fair decision trees. The main formulation of~\citet{aghaei2021} is a flow-based $\milo$ formulation where correctly classified datapoints successfully \emph{flow} through the tree; a tailored Benders' decomposition is used for large size instances.~\citet{aghaei2021} provide supporting theorems and proofs to show the strength of their formulation against OCT~\citep{bertsimas2017} and BinOCT~\citep{verwer2019} and validity of their facet defining cuts used in their Benders' decomposition.

\section{MILO Formulations}\label{milosection}
In this section, we discuss a recent mixed integer linear optimization ($\milo$) formulation proposed by~\citet{aghaei2021}. Then, we propose four $\milo$ formulations (two flow-based and two cut-based) for designing optimal binary classification trees. For brevity purposes and ease of comparisons, we employ similar notations employed by~\citet{aghaei2021}. Given a training dataset $\mathcal{T} := \{x^i,y^i\}_{i \in I}$ consisting of datapoints indexed in the set $I$. Each row $i \in I$ of $\mathcal{T}$ consists of binary features, indexed in the set $F$ and collected in the vector $x^i \in \{0,1\}^{|F|}$, and a label $y^i$, drawn from the finite set of $K$ classes. $F$ is the result of some encoding process (e.g. one-hot encoding, bucketization, binning, etc.) allowing for both discrete and continuous feature sets. For every datapoint $i \in I$ and every feature $f \in F$, binary parameter $x^i_f$ equals one if datapoint $i$ observes feature $f$. Directed graph $G_h=(V,E)$ denotes the binary decision tree with depth $h$, where $1 \le h \in \mathbb{N}$ is the maximal depth of a classification vertex in the assigned decision tree. The number of vertices and edges of $G_h$ are represented by $n := |V| = 2^{h+1} - 1$ and $m := |E| = 2^{h+1} - 2$, respectively. The vertex set $V$ is the union of the branching vertex set, $[1,\dots,2^h-1]= B \subset V$, and the leaf vertex set, $[2^h,\dots,2^{h+1}-1]=L \subset V$, with $B \cap L = \emptyset$.

\begin{figure}[H]
\centering
\begin{tikzpicture}[scale=0.52]
\begin{scope}[every node/.style={circle,thick,draw}]
\node (1) at (7.5,9) [label={[align=center,font=\fontsize{10.5}{0}\selectfont]left:$b_{11} = 1$}, label={[align=center,font=\fontsize{10.5}{0}\selectfont]right:$p_1 = 0$}]{$1$};
\node (2) at (1.5,7) [label={[align=center,font=\fontsize{10.5}{0}\selectfont]left:$b_{23} = 1$}, label={[align=center,font=\fontsize{10.5}{0}\selectfont]right:$p_2=0$}] {$2$};
\node(3) at (13.5,7) [color=blue,label={[align=center,font=\fontsize{10.5}{0}\selectfont]left:$w_{31} = 1$}, label={[align=center,font=\fontsize{10.5}{0}\selectfont]right:$p_3 = 1$}] {$3$};
\node (4) at (-2,4) [color=blue, label={[align=center,font=\fontsize{10.5}{0}\selectfont]left:$w_{42} = 1$}, label={[align=center,font=\fontsize{10.5}{0}\selectfont]right:$p_4=1$}] {$4$};
\node (5) at (5,4) [color=blue,label={[align=center,font=\fontsize{10.5}{0}\selectfont]left:$w_{53} = 1$}, label={[align=center,font=\fontsize{10.5}{0}\selectfont]right:$p_5=1$}] {$5$};
\node (6) at (10,4) [color=red, dashed, label={[align=center,font=\fontsize{10.5}{0}\selectfont]right:$p_6=0$}] {$6$};
\node (7) at (17,4) [color=red, dashed, label={[align=center,font=\fontsize{10.5}{0}\selectfont]right:$p_7=0$}] {$7$};
\end{scope}
\begin{scope}[>={Stealth[black]}]
\draw [-] (1) -- (2) node [] {};
\draw [-] (1) -- (3) node [] {};
\draw [-] (2) -- (4) node [] {};
\draw [-] (2) -- (5) node [] {};
\draw [dashed] (3) -- (6) node [] {};
\draw [dashed] (3) -- (7) node [] {};

\draw[decoration={brace,mirror,raise=5pt},decorate]
  (-6,10) -- node[left=6pt] {$B$} (-6,6);
\draw[decoration={brace,mirror,raise=5pt},decorate]
  (-6,5) -- node[left=6pt] {$L$} (-6,3);
\end{scope}
\end{tikzpicture}\hspace{8mm}
\caption{Input decision tree $G_2=(B \cup L, E)$, branching vertex set $B=\{1,2,3\}$ and leaf vertex set $L=\{4,5,6,7\}$. Here, vertices $1$ and $2$ are branched on features 1 and 3, respectively; vertices $3$, $4$, and $5$ are assigned to a classes 1, 2, and 3, respectively; and vertices 6 and 7 are pruned. 
}
\label{fig:basetree}
\end{figure}
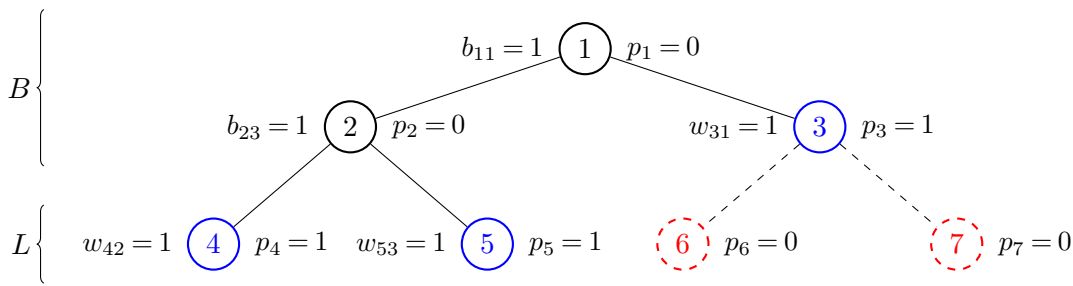

An optimal binary classification tree can be obtained by solving a biobjective optimization problem that seeks to (i) maximize the number of correctly classified datapoints and (ii) minimize the number of branching vertices. For every vertex $v \in V$, let $P_{1,v}$ and vertex set $V(P_{1,v})$ denote the unique $1,v$-path from vertex 1 to vertex $v$ and the vertex set on the path $P_{1,v}$ (including vertices $1$ and $v$), respectively. For every datapoint $i \in I$ and every vertex $v \in V$, binary variable $s^i_v$ equals one if datapoint $i$ is correctly classified at vertex $v$. For every vertex $v \in B$ and every feature $f \in F$, binary variable $b_{vf}$ equals one if vertex $v$ is branched on feature $f$. For every vertex $v \in V$ and every class $k \in K$, binary variable $w_{vk}$ equals one if vertex $v$ is assigned to class $k$. Finally for every vertex $v \in V$, binary variable $p_v$ equals one if a prediction class is assigned to vertex $v$. Figure~\ref{fig:basetree} illustrates directed graph $G_2$ with assigned decision variables $p, b,$ and $w$.

Employing binary decision variables for assigning binary tree structure is a natural approach when considering the rules proposed by~\citet{bertsimas2017}. Many current methods ($\milo$ or other) use a weighted objective function to control tree sparsity and classification simultaneously. We instead choose a bi-objective approach for a number of reasons. Without proper normalization of the classification metric, the objective function may be unfairly weighted when trying to find very sparse trees; tuning the weight parameter may also be challenging. The Pareto frontiers generated by bi-objective approaches provide efficient methods of analyzing tree sparsity on classification metrics. The frontiers are easy to interpret visualizations to compare tree out-of-sample performance with respect to sparsity and depth (or model solution metrics with respect to tree topology). Since it is known solutions of smaller trees are valid warm starts to solutions of larger trees, the frontiers are generated efficiently.

A biobjective base model for designing optimal binary classification trees is provided below.
\begin{subequations}
\label{base_model}
\begin{align}
& \max~\sum_{i \in I} \sum_{v \in V} s^i_v \label{basemax}\\
& \min~\sum_{v \in B} \sum_{f \in F} b_{vf} \label{basemin}\\
& p_v = \sum_{k \in K} w_{vk} & \forall v \in V \label{base1}\\
(\base)~~~&\sum_{f \in F} b_{vf} + \sum_{u \in V(P_{1,v})} p_u =1 & \forall v \in V \label{base2}\\
& b_{vf} = 0 & \forall v \in L,~\forall f \in F \label{base3}\\
& s^i_v \leq w_{vk} & ~\forall i \in I: y^i=k,~\forall k \in K,~\forall v \in V \label{base4}\\
& p \in [0,1]^n,~b \in \{0,1\}^{n \times |F|}, \nonumber\\
& w \in \{0,1\}^{n \times |K|},~s \in \{0,1\}^{|I| \times n}.\label{base5}
\end{align}
\end{subequations}
Here, objective function~\eqref{basemax} maximizes the number of correct classifications. Objective function~\eqref{basemin} minimizes the number of branching vertices. Constraints~\eqref{base1} imply that a vertex is labeled with a prediction class if and only if it is assigned to a class $k \in K$. Constraints~\eqref{base2} imply that every vertex $v \in V$ is either branched on a feature or a vertex on the $1,v$-path is assigned to a prediction class. Constraints~\eqref{base3} imply that no leaf vertex is branched on a feature. Constraints~\eqref{base4} imply that if datapoint $i \in I$ is classified at vertex $v \in V$, then vertex $v$ is assigned to the class for which $k = y^i$. Constraints~\eqref{base5} specify the domain of all decision variables of the base model. Furthermore, the polytope of the base model~\eqref{base_model} is denoted as follows,
\begin{align*}
    \baseP \coloneqq \{(p,b,w,s) \in [0,1]^{n(1+|I|+|F|+|K|)} : (p,b,w,s) \text{ satisfies constraints~\eqref{base1}-\eqref{base4}}\}.
\end{align*}

\begin{remark}
Constraints $p \in [0,1]^n$, $w \in \{0,1\}^{n \times |K|}$, and~\eqref{base1} imply that $p \in \{0,1\}^n$.
\end{remark}

Base model~\eqref{base_model} ensures a valid binary decision tree structure. The base model is pruning aware through constraints~\eqref{base1} to prevent over-fitting in the final decision tree; our models still produce balanced decision tree solutions. Further, we can find optimal trees when an optimal depth is not known \emph{a priori} by choosing $h$ to be large, increasing the search space and solution time. The bi-objective approach also allows for objective~\eqref{basemin} to be used as a constraint to control the sparsity of the final decision tree. However, the base model~\eqref{base_model} does not guarantee feasible connected paths for datapoints to reach a correct classification vertex; that is, a datapoint can ``jump'' onto classification vertex 5 without going through branching vertex 2 in Figure~\ref{fig:basetree}. Hence, we need complementary $\milo$ formulations with connectivity constraints to ensure that correctly classified datapoints reach their corresponding classification vertex via feasible connected paths through the tree. In the following sections, we discuss five connectivity formulations: the flow-based connectivity formulation of~\citet{aghaei2021}, two new flow-based formulations ($\mcfone$ and $\mcftwo$), and two new cut-based formulations ($\cutone$ and $\cuttwo$). These formulations are inspired by max-flow/min-cut formulations, as decision trees can be thought of as directed acyclic networks.

\subsection{Flow-based formulations}\label{flowformulations}
To impose a datapoint's feasible path through the tree, decision datapoint flow variables are defined to find a connected $1,v$-flow (path from $1$ to $v$, inclusive). Currently $G_h$ can be thought of as a directed acyclic graph with a single source, the root, and multiple sinks, leaf set $L$. Thus a datapoint \emph{flowing} through the tree could end at more than one classification vertex. To resolve this issue the tree must be augmented to have a single sink. It was~\citet{aghaei2021} who first proposed such an augmentation of the decision tree by appending a single source connected only to the root vertex and a single sink connected to every vertex in the tree.~\citet{bertsimas2017} and~\citet{hochbaum2018} both propose network flow based models but do not augment the tree. Network flow connectivity constraints are defined on this augmented tree.

Consider augmentation of input decision tree $G_h$ into a tree with a single source and sink as follows. Let $G^t_h = (V^t, A)$ with $V^t \coloneqq V \cup \{t\}$ and $A \coloneqq \{(u,v) \mid \{u,v\} \in E,~u<v\} \cup \{(v,t) \mid v \in V\}$. Directed graph $G^t_h$ adds a directed edge from $v \to t$ for all vertices of $V(G_h)$, thereby creating single sink vertex $t$. For every vertex $v \in V$, let (i) $\ell(v)$ be the left child of vertex $v$; (ii) $r(v)$ be the right child of vertex $v$; and (iii) $a(v)$ be the ancestor (parent) of vertex $v$ in arborescence $G_h$. Note that for every $v \in L$, we have $\ell(v) = r(v) = \emptyset$.

\subsubsection{FlowOCT}\label{aghasection}
\begin{figure}[H]
\centering
\begin{tikzpicture}[scale=0.3]
\begin{scope}[every node/.style={circle,thick,draw, font=\fontsize{10.5}{0}\selectfont}]
\node (0) at (8, 15.5) [] {$r$};
\node (1) at (8,9) [label={[yshift=6mm,align=center,font=\fontsize{10.5}{0}\selectfont]below:$p_1 = 0$}] {$b_{11}$};
\node (2) at (-7,7) [label={[align=center,font=\fontsize{10.5}{0}\selectfont]left:$p_2=0$}] {$b_{23}$};
\node(3) at (20,7) [color=blue, label={[font=\fontsize{10.5}{0}\selectfont]right:$p_3$}] {$w_{31}$};
\node (4) at (-14,3) [color=blue, label={[font=\fontsize{10.5}{0}\selectfont]left:$p_4$}] {$w_{42}$};
\node (5) at (0,3) [color = blue, label={[font=\fontsize{10.5}{0}\selectfont]right:$p_5$}] {$w_{51}$};
\node (6) at (16,3) [color=red, dashed, label={[align=center,font=\fontsize{10.5}{0}\selectfont]left:$p_6=0$}] {$6$};
\node (7) at (26,3) [color=red, dashed, label={[align=center,font=\fontsize{10.5}{0}\selectfont]left:$p_7=0$}] {$7$};
\node (8) at (8,0) [] {$t$};
\end{scope}
\begin{scope}[>={Stealth[black]}]
\draw [->] [-{Latex[length=2mm]}, line width=0.4mm, color = ForestGreen] (0) to [bend right = 55] node [auto=swap,midway, fill=white, font=\fontsize{10.5}{0}\selectfont] {$z^{1}_{r1}$} (1);
\draw [->] [-{Latex[length=2mm]}, line width=0.4mm, color = orange] (0) to node [auto=swap,midway, fill=white, font=\fontsize{10.5}{0}\selectfont] {$z^{2}_{r1}$} (1);
\draw [->] [-{Latex[length=2mm]}, line width=0.4mm, color = Plum] (0) to [bend left = 55] node [auto=swap,midway, fill=white, font=\fontsize{10.5}{0}\selectfont] {$z^{3}_{r1}$} (1);
\draw [->] [-{Latex[length=2mm]}, line width=0.4mm, color = ForestGreen] (1) to [bend right = 10] node [auto=swap,midway, fill=white, font=\fontsize{10.5}{0}\selectfont] {$z^{1}_{12}$} (2);
\draw [->] [-{Latex[length=2mm]}, line width=0.4mm, color = orange] (1) to [bend left = 15] node [auto=swap,midway, fill=white, font=\fontsize{10.5}{0}\selectfont] {$z^{2}_{12}$} (2);
\draw [->] [-{Latex[length=2mm]}, line width=0.4mm, color = Plum] (1) to node [auto=swap,midway, fill=white, font=\fontsize{10.5}{0}\selectfont] {$z^{3}_{13}$} (3);
\draw [->] [-{Latex[length=2mm]}, line width=0.4mm, color = ForestGreen] (2) to node [auto=swap,midway, fill=white, font=\fontsize{10.5}{0}\selectfont] {$z^{1}_{24}$} (4);
\draw [->] [-{Latex[length=2mm]}, line width=0.4mm, color = orange] (2) to node [auto=swap,midway, fill=white, font=\fontsize{10.5}{0}\selectfont] {$z^{2}_{25}$} (5);
\draw [->] [-{Latex[length=2mm]}, line width=0.4mm, dashed] (3) -- (6) node [] {};
\draw [->] [-{Latex[length=2mm]}, line width=0.4mm, dashed] (3) -- (7) node [] {};
\draw [->] [-{Latex[length=2mm]}, line width=0.4mm, color = Plum] (3) to [bend left = 45] node [auto=swap,midway, fill=white, font=\fontsize{10.5}{0}\selectfont] {$s^3_{3}$} (8);
\draw [->] [-{Latex[length=2mm]}, line width=0.4mm, color = ForestGreen] (4) to [bend right = 10] node [auto=swap,midway, fill=white, font=\fontsize{10.5}{0}\selectfont] {$s^1_{4}$} (8);
\draw [->] [-{Latex[length=2mm]}, line width=0.4mm, color = orange] (5) to [] node [auto=swap,midway, fill=white, font=\fontsize{10.5}{0}\selectfont] {$s^2_{5}$} (8);
\end{scope}
\end{tikzpicture}
\caption{Directed decision tree $G^t_2$ with decision variables of $\flowoct$. Source node $r$ is not used in our formulation of $\flowoct$ but shown for understanding datapoint flows. All decision variables shown equal one unless otherwise stated.}
\label{fig:flowoctfigure}
\end{figure}
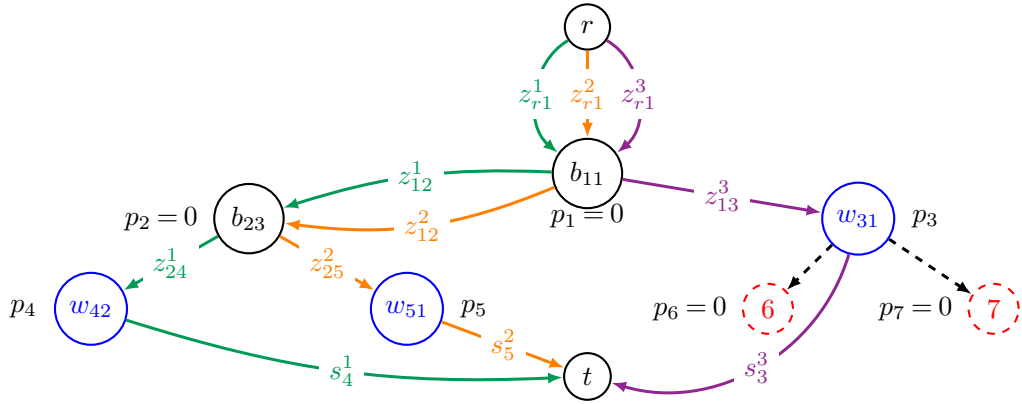

Here we present the connectivity constraints of the imbalanced tree formulation of~\citet{aghaei2021} with modifications for notation purposes. For every datapoint $i \in I$ and every directed edge $(u,v) \in A$, variable $z_{uv}^i$ denotes the flow of type $i$ on edge $(u,v)$. Since we have $\ell(v) = r(v) = \emptyset$ for $v \in L$, for every datapoint $i \in I$ and every vertex $v \in L$ we set variables $z_{v,\ell(v)}^i$ and $z_{v,r(v)}^i$ to zero. Furthermore, we do not employ the auxiliary vertex $r$, used in the original paper, in our formulation of $\flowoct$. For every datapoint $i \in I$ and every vertex $v \in V$, we also employ decision variable $s^i_v$ instead of flow variable $z_{vt}^i$ in formulation $\flowoct$ to denote the terminal vertex, $v$, of datapoint $i$. Figure~\ref{fig:flowoctfigure} illustrates directed graph $G^t_2$ and the corresponding flow variables of formulation $\flowoct$.
\begin{subequations}
\label{sinaFormulation}
\begin{align}
& (p,b,w,s) \in \baseP \label{sina0}\\
& z^i_{1,\ell(1)} + z^i_{1,r(1)}+s^i_1 \le 1 & \forall i \in I \label{sina1}\\
& z^i_{a(v),v}=z^i_{v,\ell(v)}+z^i_{v,r(v)}+s^i_v & \forall v \in V \setminus \{1\},~\forall i \in I \label{sina2}\\
(\flowoct)~~~&z^i_{v,\ell(v)} \leq \sum_{f \in F:x^i_f=0} b_{vf} \nonumber\\
& z^i_{v,r(v)} \leq \sum_{f \in F:x^i_f=1} b_{vf} & \forall v \in B,~\forall i \in I \label{sina3}\\
& b \in \{0,1\}^{n \times |F|},~w \in \{0,1\}^{n \times |K|}, \nonumber\\
& s \in \{0,1\}^{|I| \times n},~z \in \mathbb{R}^{|I| \times m}_+.\label{sina4}
\end{align}
\end{subequations}

\vspace{-4ex}

Here, constraints~\eqref{sina1} imply that for every datapoint $i \in I$ at most one unit of flow of type $i$ can emanate from vertex 1; a $\leq$ is present to allow for datapoints with no feasible flows. As we do not employ source vertex $r$, these constraints are equivalent to one unit of flow for every datapoint $i \in I$ emanates from vertex $r$. Constraints~\eqref{sina2} imply flow conservation for any datapoint $i \in I$ at any vertex $v \in V \setminus \{1\}$. Constraints~\eqref{sina3} imply that if a flow of type $i \in I$ is directed towards the left (right) child of a vertex, then the vertex is branched on a feature $f \in F$ with $x^i_f = 0$ ($x^i_f=1$).

The original formulation of $\flowoct$ augments directed graph $G_h$, which has multiple sinks, into a directed acyclic single sink graph $G^t_h$ by appending a directed edge from source node $s$ to the root vertex and directed edges from every vertex $v \in V$ to sink vertex $t$. From there well established network flow theory is used to find connected feasible paths through the tree for a datapoint. Branching is done through flow variables $z$. Successful classification of a datapoint is defined by a connected $s,t$ path through the tree.
We define the polytope of $\flowoct$ formulation as, 
\begin{align*}
    \flowoctP \coloneqq \{(p,b,w,s) \in \baseP,~z \in \mathbb{R}^{|I| \times m}_+ : (p,b,w,s,z)~\text{satisfies constraints~\eqref{sina1}-\eqref{sina3}}\}.
\end{align*}
\begin{remark}\label{flowremark}
The following constraints are implied in $\flowoctP$.
\begin{align*}
    &\sum_{v \in V} s^i_v \le 1 &\forall i \in I.
\end{align*}
\end{remark}

\subsubsection{MCF1}\label{mcf1section}
Here, we propose an extended version of formulation $\flowoct$ with same order in number of variables and constraints. We introduce a new binary variable $q^i_v$ for every datapoint $i \in I$ and every vertex $v \in V^t$. Binary variable $q^i_v$ equals one if datapoint $i$ enters vertex $v$ to reach terminal vertex $t \in V^t$. Source node $s$ is consequently removed and classification is now defined as a connected $1,t$ path through the tree as decision variables $q$ track a datapoint's path. Figure~\ref{fig:mcf1figure} illustrates directed graph $G^t_2$ and the corresponding flow variables of formulation $\mcfone$.
\begin{subequations}
\label{mcfFormulation}
\begin{align}
& (p,b,w,s) \in \baseP \label{mcf10}\\
& z_{a(v),v}^i \le  q^i_v & \forall v \in V \setminus \{1\},~\forall i \in I\label{mcf11}\\
& z_{a(v),v}^i = z_{v,\ell(v)}^i + z_{v,r(v)}^i + s^i_v & \forall v \in V \setminus \{1\},~\forall i \in I \label{mcf12}\\
(\mcfone)~~~& z^i_{1,\ell(1)} + z^i_{1,r(1)} + s^i_1 = q^i_t & \forall i \in I \label{mcf13}\\
& q^i_{\ell(v)} \le \sum_{f \in F: x^i_f = 0} b_{vf} \nonumber\\
& q^i_{r(v)} \le \sum_{f \in F: x^i_f = 1} b_{vf} & \forall v \in B,~\forall i \in I\label{mcf14}\\
& b \in \{0,1\}^{n \times |F|} ,~w \in \{0,1\}^{n \times |K|}, \nonumber\\
& s \in \{0,1\}^{|I| \times n},~q \in \{0,1\}^{|I| \times |V^t|},~z \in \mathbb{R}^{|I| \times m}_+. \label{mcf15}
\end{align}
\end{subequations}

\begin{figure}[H]
\centering
\begin{tikzpicture}[scale=0.3]
\begin{scope}[every node/.style={circle,thick,draw,font=\fontsize{10.5}{0}\selectfont}]
\node (1) at (0,11) [label={[align=center,font=\fontsize{10.5}{0}\selectfont, yshift=5mm]below:$p_1=0$}] {$b_{11}$};
\node (2) at (-15,9) [label={[align=center,font=\fontsize{10.5}{0}\selectfont,yshift=4mm]below:$p_2=0$}, label={[align=center,font=\fontsize{10.5}{0}\selectfont, yshift=8mm, xshift=4mm, color=ForestGreen]left:$q^1_2$}, label={[align=center, yshift=8mm, xshift=9mm, font=\fontsize{10.5}{0}\selectfont, color=orange]left:$q^2_2$}] {$b_{23}$};
\node(3) at (11,9)  [color=blue, label={[font=\fontsize{10.5}{0}\selectfont]above:$p_3$}, label={[font=\fontsize{10.5}{0}\selectfont, color=Plum]right:$q^3_3$}] {$w_{31}$};
\node (4) at (-24,5) [color=blue, label={[font=\fontsize{10.5}{0}\selectfont]below:$p_4$}, label={[font=\fontsize{10.5}{0}\selectfont, color=ForestGreen]left:$q^1_4$}] {$w_{42}$};
\node (5) at (-6,5) [color = blue, yshift=-4mm, label={[font=\fontsize{10.5}{0}\selectfont]left:$p_5$}, label={[font=\fontsize{10.5}{0}\selectfont, color=orange]right:$q^2_5$}] {$w_{53}$};
\node (6) at (4,5) [color=red, dashed, label={[align=center,font=\fontsize{10.5}{0}\selectfont]right:$p_6=0$}] {$6$};
\node (7) at (18,5) [color=red, dashed, label={[align=center,font=\fontsize{10.5}{0}\selectfont]left:$p_7=0$}] {$7$};
\node (8) at (0,-2) [] {$t$};
\end{scope}
\begin{scope}[>={Stealth[black]}]
\draw [->] [-{Latex[length=2mm]}, line width=0.4mm, color = orange] (1) to [bend left = 15] node [auto=swap,midway, fill=white, font=\fontsize{10.5}{0}\selectfont] {$z^{2}_{12}$} (2);
\draw [->] [-{Latex[length=2mm]}, line width=0.4mm, color = ForestGreen] (1) to [bend right = 20] node [auto=swap,midway, fill=white, font=\fontsize{10.5}{0}\selectfont] {$z^{1}_{12}$} (2);
\draw [->] [-{Latex[length=2mm]}, line width=0.4mm, color = Plum] (1) to [] node [auto=swap,midway, fill=white, font=\fontsize{10.5}{0}\selectfont] {$z^{3}_{13}$} (3);
\draw [->] [-{Latex[length=2mm]}, line width=0.4mm, color = ForestGreen] (2) to node [auto=swap,midway, fill=white, font=\fontsize{10.5}{0}\selectfont] {$z^{1}_{24}$} (4);
\draw [->] [-{Latex[length=2mm]}, line width=0.4mm, color = orange] (2) to node [auto=swap,midway, fill=white, font=\fontsize{10.5}{0}\selectfont] {$z^{2}_{25}$} (5);
\draw [->] [-{Latex[length=2mm]}, line width=0.4mm, dashed] (3) -- (6) node [] {};
\draw [->] [-{Latex[length=2mm]}, line width=0.4mm, dashed] (3) -- (7) node [] {};
\draw [->] [-{Latex[length=2mm]}, line width=0.4mm, color = Plum] (3) to [bend left = 40] node [auto=swap,midway, fill=white, font=\fontsize{10.5}{0}\selectfont] {$s^3_{3}$} (8);
\draw [->] [-{Latex[length=2mm]}, line width=0.4mm, color = ForestGreen] (4) to [bend right = 10] node [auto=swap,midway, fill=white, font=\fontsize{10.5}{0}\selectfont] {$s^1_{4}$} (8);
\draw [->] [-{Latex[length=2mm]}, line width=0.4mm, color = orange] (5) to [] node [auto=swap,midway, fill=white, font=\fontsize{10.5}{0}\selectfont] {$s^2_{5}$} (8);
\end{scope}
\end{tikzpicture}
\caption{Directed decision tree $G^t_2$ with decision variables of $\mcfone$. All decision variables shown equal one unless otherwise stated.}
\label{fig:mcf1figure}
\end{figure}

Here, constraints~\eqref{mcf11} and~\eqref{mcf12} imply that if vertex $v \in V^t$ receives a flow of type datapoint $i \in I$, then vertex $v$ is selected on a $1,t$ path of datapoint $i$. Constraints~\eqref{mcf12} have equivalent interpretations to constraints~\eqref{sina2}. Constraints~\eqref{mcf13} imply that if datapoint $i \in I$ reaches vertex $t$, then either the datapoint is correctly classified at vertex $1$ or a flow of type datapoint $i$ is generated at vertex 1. Constraints~\eqref{mcf14} imply that if the left (right) child of parent $v \in V$ is selected on the path of datapoint $i \in I$, then $x^i_f = 0$ ($x^i_f = 1$). Constraints~\eqref{mcf14} are analogous to constraints~\eqref{sina3}; in $\mcfone$ datapoint left-right branching is controlled by binary variables $q$ whereas in $\flowoct$ flow variables $z$ control datapoint left-right branching. 

We define an extended model of $\base$ called $\baseq$ to be objective functions~\eqref{basemax} and~\eqref{basemin} with constraints~\eqref{base1}-\eqref{base5},~\eqref{mcf14} and~\eqref{mcf15}, and its poltyope, $\baseqP$, as
\begin{align*}
    \baseqP \coloneqq \{&(p,b,w,s,q) \in [0,1]^{n(1+|F|+|K|+2|I|)}:(p,b,w,s,q) \text{satisfies~\eqref{base1}-\eqref{base4} and~\eqref{mcf14}}\}.
\end{align*}
We then define the polytope of the $\mcfone$ model as,
\begin{align*}
    \mcfoneP \coloneqq \{&(p,b,w,s,q) \in \baseqP,~q \in[0,1]^{|I|\times|V^t|},~z \in \mathbb{R}^{|I| \times m}_+:(p,b,w,s,q,z)~\text{satisfies~\eqref{mcf11}-\eqref{mcf13}} \}.
\end{align*}
\begin{remark}\label{mcf1remark}
The following constraints are implied in $\mcfoneP$.
\begin{align*}
    &\sum_{v \in V} s^i_v =  q^i_t &\forall i \in I.
\end{align*}\label{implied_mcf1_ineq}
\end{remark}\vspace{-6ex}

\subsubsection{MCF2}\label{mcf2section}
This is a multi-commodity flow formulation in which flows are labeled by datapoints and their destination. No augmentation of the graph is required and we use $G_h$. For every datapoint $i \in I$, every vertex $v \in V \setminus \{1\}$, and every directed edge $(c,d) \in E$, variable $z_{cd}^{iv}$ denotes the flow of type datapoint $i$, emanating from the root and terminating at $v$ on directed edge $(c,d) \in E$. Successful classification through the tree is now a connected $1,v$-path. Figure~\ref{fig:mcf2figure} illustrates directed graph $G_2$ and the corresponding flow variables of formulation $\mcftwo$.
\begin{subequations}
\label{formulationMCF2}
\begin{align}
& (p,b,w,s,q) \in \baseqP \label{mcf20}\\
& z_{1,\ell(1)}^{iv} + z_{1,r(1)}^{iv} = s^i_v & \forall v \in V \setminus \{1\},~\forall i \in I \label{mcf21}\\
& z^{iv}_{u,\ell(u)} + z^{iv}_{u,r(u)} - z^{iv}_{a(u),u} = 0 & \forall u \in V \setminus \{1,v\},~\forall v \in V \setminus \{1\},~\forall i \in I \label{mcf22}\\
& \sum_{u \in V \setminus \{1\}} z^{iu}_{a(v),v} \le q^i_v & \forall v \in V \setminus \{1\},~\forall i \in I \label{mcf23}\\
& z^{iv}_{a(v),v} = s^i_v & \forall v \in V \setminus \{1\},~\forall i \in I \label{mcf23.5}\\
& \sum_{v \in V} s^i_v \le 1 & \forall i \in I\label{mcf25}\\
& b \in \{0,1\}^{n \times |F|},~w \in \{0,1\}^{n \times |K|}, \nonumber\\
& s \in \{0,1\}^{|I| \times n},~q \in \{0,1\}^{|I| \times n}, \nonumber\\
& z \in \mathbb{R}^{|I| \times n \times m }_+.\label{mcf26}
\end{align}
\end{subequations}

\begin{figure}[H]
\centering
\begin{tikzpicture}[scale=0.3]
\begin{scope}[every node/.style={circle,thick,draw,font=\fontsize{10.5}{0}\selectfont}]
\node (1) at (6,8) [label={[align=center,font=\fontsize{10.5}{0}\selectfont,yshift=5mm]below:$p_1=0$}] {$b_{11}$};
\node (2) at (-8.5,4) [label={[align=center,yshift=5mm,font=\fontsize{10.5}{0}\selectfont]below:$p_2=0$}, label={[align=center,font=\fontsize{10.5}{0}\selectfont, xshift=2.5mm, yshift=7.5mm, color=ForestGreen]left:$q^1_2$}, label={[align=center,font=\fontsize{10.5}{0}\selectfont, yshift=7.5mm, xshift=7.5mm, color=orange]left:$q^2_2$}] {$b_{23}$};
\node(3) at (20.5,4) [color=blue, label={[font=\fontsize{10.5}{0}\selectfont]left:$p_3$}, label={[font=\fontsize{10.5}{0}\selectfont, yshift=6mm, color=Plum]right:$q^3_3$}, label={[font=\fontsize{10.5}{0}\selectfont, color=Plum]right:$s^3_3$}] {$w_{31}$};
\node (4) at (-17,0) [color=blue, label={[font=\fontsize{10.5}{0}\selectfont]right:$p_4$}, label={[font=\fontsize{10.5}{0}\selectfont, yshift=3.5mm, color=ForestGreen]left:$q^1_4$}, label={[font=\fontsize{10.5}{0}\selectfont, yshift=-2.5mm, color=ForestGreen]left:$s^1_4$}] {$w_{42}$};
\node (5) at (0,0) [color=blue, label={[font=\fontsize{10.5}{0}\selectfont]left:$p_5$}, label={[font=\fontsize{10.5}{0}\selectfont, yshift=3.5mm, color=orange]right:$q^2_5$}, label={[font=\fontsize{10.5}{0}\selectfont, yshift=-2.5mm, color=orange]right:$s^2_5$}] {$w_{53}$};
\node (6) at (12,0) [color=red, dashed, label={[align=center,font=\fontsize{10.5}{0}\selectfont]right:$p_6=0$}] {$6$};
\node (7) at (28,0) [color=red, dashed, label={[align=center,font=\fontsize{10.5}{0}\selectfont]left:$p_7=0$}] {$7$};
\end{scope}
\begin{scope}[>={Stealth[black]}]
\draw [->] [-{Latex[length=2mm]}, line width=0.4mm, color = orange] (1) to [bend left = 15] node [auto=swap,midway, fill=white, font=\fontsize{10.5}{0}\selectfont] {$z^{25}_{12}$} (2);
\draw [->] [-{Latex[length=2mm]}, line width=0.4mm, color = ForestGreen] (1) to [bend right = 20] node [auto=swap,midway, fill=white, font=\fontsize{10.5}{0}\selectfont] {$z^{14}_{12}$} (2);
\draw [->] [-{Latex[length=2mm]}, line width=0.4mm, color = Plum] (1) to node [auto=swap,midway, fill=white, font=\fontsize{10.5}{0}\selectfont] {$z^{33}_{13}$} (3);
\draw [->] [-{Latex[length=2mm]}, line width=0.4mm, color = ForestGreen] (2) to node [auto=swap,midway, fill=white, font=\fontsize{10.5}{0}\selectfont] {$z^{14}_{24}$} (4);
\draw [->] [-{Latex[length=2mm]}, line width=0.4mm, color = orange] (2) to node [auto=swap,midway, fill=white, font=\fontsize{10.5}{0}\selectfont] {$z^{25}_{25}$} (5);
\draw [->] [-{Latex[length=2mm]}, line width=0.4mm, dashed] (3) -- (6) node [] {};
\draw [->] [-{Latex[length=2mm]}, line width=0.4mm, dashed] (3) -- (7) node [] {};
\end{scope}
\end{tikzpicture}
\caption{Directed decision tree $G_2$ with decision variables of $\mcftwo$. All decision variables shown equal one unless otherwise stated.}
\label{fig:mcf2figure}
\end{figure}

Here, constraints~\eqref{mcf21} imply that if datapoint $i \in I$ is correctly classified at vertex $v \in V \setminus \{1\}$, then a flow of type $iv$ originates from vertex 1. Constraints~\eqref{mcf22} imply the flow conservation of datapoint $i$ heading towards vertex $v$ at vertex $u \in V \setminus \{1,v\}$. Constraints~\eqref{mcf23} imply that if a flow of datapoint $i \in I$ that is heading toward vertex $u$ enters vertex $v$, then vertex $v$ is selected on the $1,u$-path. Constraints~\eqref{mcf23.5} imply that datapoint $i \in I$ is correctly classified at vertex $v \in V \setminus \{1\}$ if and only if a flow of type $iv$ enters vertex $v$. 
Constraints~\eqref{mcf25} imply that each datapoint $i \in I$ can be correctly classified in at most one vertex. We define the polytope of the $\mcftwo$ model as,
\begin{align*}
    \mcftwoP \coloneqq & \{(p,b,w,s,q) \in \baseqP,~z \in \mathbb{R}^{|I| \times n \times m}_+ :(p,b,w,s,q,z)~\text{satisfies constraints~\eqref{mcf21}-\eqref{mcf25}} \}
\end{align*}

\begin{remark}\label{mcf2remark}
The following constraints are implied in $\mcftwoP$.
\begin{align}
    &z^{iv}_{a(v),v} - z^{iv}_{v, \ell(v)} - z^{iv}_{v, r(v)} = s^i_v &\forall v \in V \setminus \{1\},~\forall i \in I,\label{remark31}\\
    &z^{iv}_{v, \ell(v)} + z^{iv}_{v, r(v)} = 0 & \forall v \in V \setminus \{1\},~\forall i \in I\label{remark32}.
\end{align}
\end{remark}
Proofs of Remarks~\ref{flowremark}~-~\ref{mcf2remark} are found in Appendix~\ref{appendix:remarks}.

\subsection{Cut-based formulations}\label{cutformulations}
In this section, we propose two cut-based formulations, both of which have connectivity constraints that are added on-the-fly. Further, their corresponding separation problems are solved in polynomial time; the $1,v$-path of any vertex $v \in V(G_h)$ is found in $\mathcal{O}(|V|)$~\citep{kaplan2011} as a tree itself is a directed acyclic graph. The motivation behind our cut based formulation is the $P_{1,v}$ of any vertex $v \in V$ is unique since $G_h$ is a tree. Thus any vertex $c \in V(P_{1,v})$ is a valid $1,v$-separator. Through our definition of variables $q$ and $s$ we can find $1,v$-separators for a terminal vertex of a datapoint $i \in I$ to find a feasible path through the tree.

\subsubsection{CUT1}
This formulation is the reduced formulation of $\mcftwo$ as we remove flow variables $z$. Here we aim to find $1,v$-separators strictly on the $1,v$-path. For every vertex $v \in V$, we define $P_v$ as the set of non-root vertices on the $1,v$-path.
\begin{subequations}
\label{formulationCut1}
\begin{align}
& (p,b,w,s,q) \in \baseqP \label{cut10}\\
& \sum_{v \in V} s^i_v \le 1 & \forall i \in I\label{cut12}\\
(\cutone)~~~& s^i_v \le  q^i_c &\forall c \in V(P_v),~\forall v \in V \setminus \{1\},~\forall i \in I \label{cut13}\\
& b \in \{0,1\}^{n \times |F|} ,~w \in \{0,1\}^{n \times |K|}, \nonumber\\
& s \in \{0,1\}^{|I| \times n},~q \in \{0,1\}^{|I| \times n}. \label{cut14}
\end{align}
\end{subequations}
Here, constraints~\eqref{cut13} imply that if a datapoint $i \in I$ is classified at vertex $v \in V \setminus \{1\}$, then all vertices on the path from $1$ to $v$, excluding vertex 1, must be selected. 

We define the polytope of formulation~\eqref{formulationCut1} as,
\begin{align*}
    \cutoneP \coloneqq \{(p,b,w,s,q) \in \baseqP : (p,b,w,s,q)~\text{satisfies constraints~\eqref{cut12} and \eqref{cut13}}\}. 
\end{align*}

\subsubsection{CUT2}
We propose another cut-based formulation whose linear optimization relaxation is stronger than that of formulation $\cutone$ by redefining a $1,v$-separator as any vertex that separates terminal vertex $v$ or any one of its children $(\child(v))$. For every vertex $v \in V \setminus \{1\}$, we define
\begin{align*}
    \child(v) \coloneqq \{u \in V \setminus \{v\} : u>v,~\dist_{G_h}(v,u) < \infty \}.
\end{align*}
Here $\dist_{G_h}(v,u)$ denotes the distance between vertices $v$ and $u$ in directed graph $G_h$. For example in Figure~\ref{fig:mcf2figure}, $\child(2) = \{4,5\}$. By redefining the set of $1,v$-separators of a terminal vertex $v \in V$ we provide stronger lower bounds on decision variables $q$ when adding cuts at points in the branch and bound tree, integral or fractional. This holds as a datapoint $i \in I$ must pass through $v$ to select $v$ or any one of its children as its terminal vertex.
\begin{subequations}
\label{formulationCut2}
\begin{align}
& (p,b,w,s,q) \in \baseqP \label{cut20}\\
& \sum_{v \in V} s^i_v \le 1 & \forall i \in I\label{cut22}\\
(\cuttwo)~~~& s^i_v + \sum_{u \in \child(v)} s^i_u\le  q^i_c &\forall c \in V(P_v),~\forall v \in V \setminus \{1\},~\forall i \in I \label{cut23}\\
& b \in \{0,1\}^{n \times |F|} ,~w \in \{0,1\}^{n \times |K|}, \nonumber\\
& s \in \{0,1\}^{|I| \times n},~q \in \{0,1\}^{|I| \times n}. \label{cut24}
\end{align}
\end{subequations}
Here, constraints~\eqref{cut23} imply that if a datapoint $i \in I$ is correctly classified at vertex $v$ or one of its descendants, then the datapoint selects every vertex on the path from $1$ to $v$ excluding 1. 

We define the polytope of formulation~\eqref{formulationCut2} as,
\begin{align*}
    \cuttwoP \coloneqq \{(p,b,w,s,q) \in \baseqP : (p,b,w,s,q)~\text{satisfies constraints~\eqref{cut22} and \eqref{cut23}}\}. 
\end{align*}

\section{Theoretical comparison of formulations}\label{theorycomp}
In this section, we prove Theorem~\ref{strengththeorem}, comparing the strength of the formulations discussed in Section~\ref{milosection}. The theorem follows directly by Lemmata~\ref{cut2lemma} through~\ref{mcf1lemma}, which show formulations $\cuttwo$ through $\mcfone$ are just as strong as that of $\flowoct$. Lemma~\ref{cut2lemma} shows a strict inclusion between $\cuttwo$ and $\cutone$ through an example. Lemma~\ref{cut1lemma} shows the equivalence of formulations $\mcftwo$ and $\cutone$ due to max-flow min-cut result of~\citet{ford1962}. Lemma~\ref{mcf2lemma} shows $\mcfone$ and $\mcftwo$ are equivalent to each other due to the unique $1,v$-paths of a decision tree. Lemma~\ref{mcf1lemma} shows the strong LO relaxation of $\mcfone$ against $\flowoct$. Note that our Theorem~\ref{strengththeorem} and Theorem 1 of~\citet{aghaei2021} suffice to show our formulations are at least as strong as that of recent $\milo$ formulations OCT~\citep{bertsimas2007} and BinOCT~\citep{verwer2019}. Lastly, we show the correctness of our proposed formulations through Theorems~\ref{flowcorrect} and~\ref{cutcorrect}. For brevity purposes we provide the full proofs for Theorem~\ref{strengththeorem} and Lemmata~\ref{cut2lemma},~\ref{mcf1lemma} in Section~\ref{theorycomp} and the full proofs for Lemmata~\ref{cut1lemma},~\ref{mcf2lemma} and Theorems and \ref{flowcorrect},~\ref{cutcorrect} in Appendices~\ref{appendix:lemmas} and~\ref{appendix:theorems}, respectively.

\begin{theorem}[Strength of Formulations]\label{strengththeorem}
Let $X=(p,b,w,s,q)$. Then,
\begin{align*}
\cuttwoP \subseteq \cutoneP = \proj_{X}\mcftwoP = \proj_{X}\mcfoneP \subseteq \proj_{X}\flowoctP.
\end{align*}
\end{theorem}

\proof{Proof of Theorem~\ref{strengththeorem}.}\label{strengthproof}
The proof follows by Lemmata~\eqref{cut2lemma},~\eqref{cut1lemma},~\eqref{mcf2lemma}, and~\eqref{mcf1lemma}.
\endproof


\begin{lemma}\label{cut2lemma}
$\cuttwoP \subseteq \cutoneP$, and the inclusion can be strict. 
\end{lemma}

\proof{Proof of Lemma~\ref{cut2lemma}.}\label{cut2proof}
Let $(\hat{p}, \hat{s}, \hat{b}, \hat{w}, \hat{q})$ be a point that belongs to $\cuttwo$. We are to show that it also belongs to $\cutone$. It is clear that the point satisfies constraints~\eqref{cut10} and \eqref{cut12}. For every datapoint $i \in I$ and for every vertex $v \in V$ and for every separating vertex $c \in P_v$, we show that point $(\hat{p}, \hat{s}, \hat{b}, \hat{w}, \hat{q})$ satisfies constraints~\eqref{cut13}. 
\begin{align*}
    \hat{s}^i_v \leq \hat{s}^i_v + \sum_{u \in \child(v)} \hat{s}^i_u \leq \hat{q}^i_c
\end{align*}
The first inequality holds by the nonnegatviity of $\hat{s}^i_v$ for all $v \in V$ and $i \in I$. The second inequality holds by constraints~\eqref{cut23}. This completes the proof.

Figure~\ref{fig:badPointOne} illustrates an instance of the optimal binary decision tree problem that belongs $\cutoneP$ but not $\cuttwoP$. Here, one datapoint has three features and one class. All features equal to zero, $x^1_1 = x^1_2 = x^1_3 = 0$, and class $y^1 = 1$. The fractional point in the branch and bound tree indicates three branching vertices 1, 2, and 4, (using features 1, 2, and 3, respectively) and four classification vertices 3, 5, 8 and 9 (all using class $k=1$). The datapoint enters vertices 1, 2, 4, 8 and 9 and selects vertices 4, 8, and 9 as terminal vertices. While the fractional point is in $\cutoneP$, it does not belong to $\cuttwoP$ because it violates constraints~\eqref{cut23} for terminal vertex $v = 4$ and its separating vertex $c = 2$. In simpler terms,
\begin{align*}
    \bar{s}^1_4 + \sum_{u \in \child[4]} \bar{s}^1_u = \bar{s}_4^1 + \bar{s}_8^1 + \bar{s}_9^1 = 0.1 + 0.1 + 0 = 0.2 \not \le 0.15 = q_2^1.
\end{align*}
\endproof
\vspace{-14ex}

\begin{figure}[H]
\centering
\begin{tikzpicture}[scale=0.2]
\begin{scope}[every node/.style={circle,thick,draw}]
\node (1) at (8,12) [label={[yshift=-17mm,align=center,font=\fontsize{9}{0}\selectfont]above:$\bar{b}_{11} = 0.75,~\bar{q}_1^1 = 1$\\$\bar{p}_{1} = 0.25,~\bar{w}_{11} = 0.25$,~$\bar{s}^1_1 = 0$}] {$1$};
\node (2) at (1,8) [label={[yshift=6mm,xshift=10mm,align=center,font=\fontsize{9}{0}\selectfont]left:$\bar{b}_{22} = 0.25,~\bar{w}_{21} = 0.5$\\$\bar{p}_{2} = 0.5,~\bar{q}_2^1 = 0.15$\\$\bar{s}^1_2 = 0$}] {$2$};
\node(3) at (15,8) [label={[align=center,yshift=2mm,font=\fontsize{9}{0}\selectfont]right:$\bar{b}_{31} = \bar{b}_{32} = \bar{b}_{33} = 0$\\
$\bar{p}_{3} = 0.75,~\bar{w}_{31} = 0.75$\\$\bar{q}_3^1 = 0,~\bar{s}^1_3 = 0$}] {$3$};
\node (4) at (-6,4) [label={[yshift=3mm,xshift=4mm,align=center,font=\fontsize{9}{0}\selectfont]left:$\bar{b}_{43} = 0.1,~\bar{w}_{41} = 0.15$\\
$\bar{p}_{4} = 0.15,~\bar{q}_4^1 = 0.1$\\$\bar{s}^1_4 = 0.1$}] {$4$};
\node (5) at (8,4) [label={[yshift=-2mm,align=center,font=\fontsize{9}{0}\selectfont]right:$\bar{b}_{51} = \bar{b}_{52} = \bar{b}_{53} = 0$\\
$\bar{p}_5 = 0.25,~\bar{w}_{51} = 0.25$\\$\bar{q}_5^1 = 0,~\bar{s}^1_5 = 0$}] {$5$};
\node (8) at (-13,0) [label={[yshift=-1mm,align=center,font=\fontsize{9}{0}\selectfont]left:$\bar{b}_{81} = \bar{b}_{82} = \bar{b}_{83} = 0$\\
$\bar{p}_8 = 0.1,~\bar{w}_{81} = 0.1$\\$\bar{q}_8^1 = 0.1,~\bar{s}^1_8 = 0.1$}] {$8$};
\node (9) at (1,0) [label={[yshift=-4mm,xshift=-1mm,align=center,font=\fontsize{9}{0}\selectfont]right:$\bar{b}_{91} = \bar{b}_{92} = \bar{b}_{93} = 0$\\
$\bar{p}_9 = 0.1,~\bar{w}_{91} = 0.1$\\$\bar{q}_9^1 = 0,~\bar{s}^1_9 = 0$}] {$9$};
\end{scope}
\begin{scope}[>={Stealth[black]}]
\draw [->] [-{Latex[length=2mm]}, line width=0.4mm] (1) to [] node [] {} (2);
\draw [->] [-{Latex[length=2mm]}, line width=0.4mm] (1) to node [] {} (3);
\draw [->] [-{Latex[length=2mm]}, line width=0.4mm] (2) to node [] {} (4);
\draw [->] [-{Latex[length=2mm]}, line width=0.4mm] (2) to node [] {} (5);
\draw [->] [-{Latex[length=2mm]}, line width=0.4mm] (4) -- (8) node [] {};
\draw [->] [-{Latex[length=2mm]}, line width=0.4mm] (4) -- (9) node [] {};
\end{scope}
\end{tikzpicture}
\vspace{-6ex}
\caption{A point that belongs to $\cutoneP$, but not $\cuttwoP$.\label{fig:badPointOne}}
\end{figure}

\begin{lemma}\label{cut1lemma}
Let $X=(p,b,w,s,q)$. Then, $\cutoneP = \proj_{X}\mcftwoP$.
\end{lemma}
\begin{lemma}\label{mcf2lemma}
Let $X=(p,b,w,s,q)$. Then $\proj_{X}\mcftwoP = \proj_X\mcfoneP$.
\end{lemma}
\begin{lemma}\label{mcf1lemma}
    Let $X=(p,b,w,s,q)$. Then $\proj_X \mcfoneP \subseteq \proj_{X}\flowoctP$.
\end{lemma}

\proof{Proof of Lemma~\ref{mcf1lemma}.}\label{mcf1proof}
Let $(\hat{p}, \hat{s}, \hat{b} , \hat{w}, \hat{q}, \hat{z})$ be a point that belongs to the $\mcfoneP$ polytope. We are to show that the point belongs to the polytope $\flowoctP$. First, it is clear that $(\hat{p}, \hat{s}, \hat{b} , \hat{w}, \hat{q}, \hat{z}) \in \baseP$; so, it satisfies constraint~\eqref{sina0}.

It suffices to show that $(\hat{p}, \hat{s}, \hat{b} , \hat{w}, \hat{q}, \hat{z})$ satisfies constraints~\eqref{sina1}-\eqref{sina3}. For every datapoint $i \in I$,
\begin{align*}
    \hat{z}^i_{1,\ell(1)} + \hat{z}^i_{1,r(1)} + \hat{s}^i_1 = \hat{q}^i_t \le 1.
\end{align*}
Here, the equality holds by constraints~\eqref{mcf13}. The inequality holds because $\hat{q} \in [0,1]^{|I| \times |V^t|}$. Hence, point $(\hat{p}, \hat{s}, \hat{b} , \hat{w}, \hat{q}, \hat{z})$ satisfies constraints~\eqref{sina1}. Because $(\hat{p}, \hat{s}, \hat{b} , \hat{w}, \hat{z}, \hat{q})$ satisfies constraints~\eqref{mcf12}, the point satisfies constraints~\eqref{sina2}. Finally for every datapoint $i \in I$ and every vertex $v \in B$, we have
\begin{subequations}
\begin{align}
    \hat{z}^i_{v,\ell(v)} \le \hat{q}^i_{\ell(v)} \le \sum_{f \in F: x^i_f = 0} \hat{b}_{vf},\label{first_ineq_lemma1}\\
    \hat{z}^i_{v,r(v)} \le \hat{q}^i_{r(v)} \le \sum_{f \in F: x^i_f = 1} \hat{b}_{vf}.\label{second_ineq_lemma1}
\end{align}
\end{subequations}
Here, the first inequalities in~\eqref{first_ineq_lemma1} and~\eqref{second_ineq_lemma1} hold by constraints~\eqref{mcf11}. The second inequalities in~\eqref{first_ineq_lemma1} and~\eqref{second_ineq_lemma1} hold by constraints~\eqref{mcf14}. Hence, point $(\hat{p}, \hat{s}, \hat{b} , \hat{w}, \hat{z}, \hat{q})$ satisfies constraints~\eqref{sina3}. This concludes the proof.
\endproof
The following two theorems show the correctness of our proposed models.
\begin{theorem}\label{flowcorrect}
    Let $p^* \in \flowoctP$ be a feasible solution of formulation~\eqref{sinaFormulation}. Then $p^*$ represents a feasible binary classification tree.
\end{theorem}
\begin{theorem}\label{cutcorrect}
    Suppose a point $p^*$ is a feasible binary classification tree. Then, $p^* \in \cuttwoP$.
\end{theorem}

\section{Total Unimodularity (TU) Properties}\label{tu}
To show that $\baseq$ is TU we use an equitable bicoloring of its columns and the results of~\citet{ghoulia1962} who showed that a $0,\pm1$ matrix is totally unimodular if and only if every column submatrix has an equitable bicoloring. Matrix $A$ has an equitable bicoloring if its columns can be partitioned into red and blue columns so that, for every row of $A$, the sum of the entries in columns labeled red differs by at most one from the sum of the entries in columns labeled blue. 

Below is the LHS of the standard form representation of model $\baseq$.~$\mathit{I}_n$ is the $n$-dimensional identity matrix, $A_n \in \mathbb{R}^{n\times n}(A_{a\times b} \in\mathbb{R}^{a\times b})$, \texttt{BD}$(A)_k$ is a $k\times k$ block diagonal matrix containing the submatrix $A$ along the diagonal and 0's everywhere else, and $J_{a\times b}$ is the $a\times b$ unit matrix. Here, $B=\mathit{J}_{1\times|F|}$, $W_1=-\mathit{J}_{1\times|K|}$, and $W_2\in\mathbb{R}^{|\mathcal{T}|\times|K|}$ is the encoded matrix representation of training datapoint classes $y$. $\bar{P} \in \{0,1\}^{n\times n}$, has row and columns indexed by $v$. Each row, corresponding to decision variable $p_v$, contains a 1 in the columns of $V(P_{1,v})$ and 0's everywhere else. Lastly,
\begin{equation*}
    D = 
    \begin{bmatrix}
    \hat{\mathcal{T}} - \mathit{J}_{|\mathcal{T}|\times|F|} \\
    -\hat{\mathcal{T}}
    \end{bmatrix} \in \mathbb{R}^{2|I|\times|F|},~\text{where}~\hat{\mathcal{T}} = \mathcal{T}\setminus y,~\text{the training feature set}.
\end{equation*}
\[\small
\left[\arraycolsep=4pt\def\arraystretch{3}
\begin{array}{c:c:c:c:c}

\textbf{\huge$\mathit{I}_n$} & \LARGE{\texttt{BD}\textbf{$(W_1)_n$}} & \textbf{\huge0} & \textbf{\huge0} & \textbf{\huge0} \\\hdashline

\textbf{\huge$\bar{P}$} & \textbf{\huge0} & \LARGE{\texttt{BD}\textbf{$(B)_n$}} & \textbf{\huge0} & \textbf{\huge0} \\\hdashline

\textbf{\huge0} & \textbf{\huge0} &
\left[\def\arraystretch{2.5}\begin{array}{c}
\textbf{\huge0\Large$_{|F||B|\times|F||L|}$} \\
\textbf{\huge$\mathit{I}$\Large$_{|F||L|}$}
\end{array}\right]^\textbf{\Large$\intercal$}
& \textbf{\huge0} & \textbf{\huge0} \\\hdashline

\textbf{\huge0} & \LARGE{\texttt{BD}\textbf{$(-W_2)_n$}} & \textbf{\huge0} & \textbf{\huge$\mathit{I}$\Large$_{n|\mathcal{T}|}$} & \textbf{\huge0} \\\hdashline

\textbf{\huge0} & \textbf{\huge0} & 
\left[\def\arraystretch{2.5}\begin{array}{c}\LARGE{\texttt{BD}\textbf{$(D^\intercal)_{|B|}$}} \\
\textbf{\huge0\Large$_{|F||L|\times 2|\mathcal{T}||B|}$}
\end{array}\right]^\textbf{\Large$\intercal$}
& \textbf{\huge0} & \textbf{\huge$\mathit{I}$\large$_{2|\mathcal{T}||B|}$}
\end{array}\right]
\left[\arraycolsep=.01pt
\color{blue}\begin{array}{c}
x
\end{array}\right]
\]

where $x^\intercal = [p_1,\dots,p_n,~w_{1,1},w_{1,2},\dots,w_{n,|K|},~b_{1,1},b_{1,2},\dots,b_{n,|F|},~s^1_1,s^2_1,\dots,s^{|I|}_n,~q^1_1,q^2_1,\dots,q^{|I|}_n]$
\subsection{Equitable Bicoloring Rules}
We can label the columns of our matrix in the following manner. Let $\stackrel{L}{=}$ represent the label function.
\begin{itemize}
    \item Columns associated with decision variables $p$ are assigned $red$ or $blue$ based on the depth of vertex $v$ of variable $p_v$. Columns of vertices with even depth are labeled $red$ and odd depth columns are labeled $blue$. The root is defined to have depth 0.

    \item Columns associated with decision variables $w$ are grouped sets of size $|K|$, each set corresponding to a $v \in V$. The depth of $v$ will determine the labeling scheme. If the depth of $v$ is even we label each column in the set alternating starting with $red$, and if depth of $v$ is odd alternating starting with $blue$.

    \item Columns associated with decision variables $s$ are grouped into sets of size $|\mathcal{T}|$, each set corresponding to a $v \in V$. In a set, each column is assigned the same color as the column of $w_{vk=y^i}$, i.e. the column is labeled the same color corresponding to the datapoint's class at said vertex.

    \item Columns associated with decision variables $b$, are grouped into sets of size $|F|$, each set corresponding to a $v \in V$. Since, $\hat{\mathcal{T}} \in \{0,1\}^{|F|}$ we can safely assume that $F$ was the result of some encoding process, of which we control. Thus we know the original feature space, call it $H$. For each original feature $h_i \in H$, there exists a set $F_i = \{f_k,f_{k+1},\dots,f_n\}\subset F$ generated from a mapping, $\Omega:H\rightarrow F$, defined by the encoding procedure (i.e. unique mapping of singular original features to sets of encoded features). For each vertex $v \in V$ we follow the coloring procedure below to assign colors to the entire feature set $F$ associated with $v$.
    \begin{enumerate}
        \item Find a pair of $h \in H$, call them $h_1$ and $h_2$, that have the same number of encoded features (i.e. $|F_1|=|F_2|$). Label all the encoded features of $h_1 \coloneqq F_1~blue$, and the encoded features of $h_2 \coloneqq F_2~red$. Set $H = H\setminus\{h_1,h_2\}$. Repeat the process until no such $h_1, h_2$ exist.
        \item If there still exist $h \in H$. Construct a restricted bipartition of $H$, a bipartition that does not necessarily include all of $H$. Call it $C=\{h_j,\dots,h_k\}$ and $D=\{h_x,\dots,h_y\},~C \cap D = \emptyset$. Further take $C$ and $D$ such that $|\cup_{i=j}^k \Omega(h_i)-\cup_{i=x}^y \Omega(h_i)|\leq 1$, the sum of encoded features in each partition are equal or differ by 1. Assume that $|C| \geq |D|$. Label all mapped encoded features of $C~blue$ and mapped encoded features of $D~red$. (Note this process can be done greedily, increasing the size of the restricted-bipartitions). $H \coloneqq H \setminus \{C,D\}$.
        \item[$\ast$] It may be conducive to \emph{not} find all equal pairs of step 1 to yield a favorable restricted bipartition in step 2.
        \item If there still exists $h \in H$, there will be exactly one. Label the encoded features of $h$ alternating starting with $blue$.
    \end{enumerate}

    \item Columns associated with decision variables $q$ are categorized into two types through the definition of left, right children of vertices used for decision variables $q$. The rows of $\baseq$ can also be categorized into two types according to this left, right relationship. Due to the identity matrix corresponding to the columns of decision variables $q$, our labeling scheme uses rows types to assign labels. Each row type is size $|\mathcal{T}||B|$, covering all decision variables $q$.
    \begin{enumerate}
        \item Rows of the first type correspond to first set of constraints of~\eqref{mcf14}, branching to the left child of a vertex. For each row, observe the only non-zero entries are associated with $|F|$ columns of decision variables $b$ and exactly one from decision variables $q$. By design the columns of $b$ admit an equitable bicoloring value of 0 or 1, favoring $red$. Thus all columns all of $q$ for the first type of rows are labeled $blue$ to ensure the entire row admits an equitable bicoloring.
        \item Each row of the second type, corresponding to branching to the right child of a vertex, will have $|F|$ -1's from columns of $b$. If the equitable bicoloring value for the columns of $b$ is 0, label the non-zero column of $q$ $blue$. If the value is -1, label the non-zero column of $q$ the same color the bicoloring favors.
    \end{enumerate}
\end{itemize}
For examples of the labeling procedure refer to Table~\ref{tab:bicoloringexamples} in Appendix~\ref{appendix:bicoloring}.

\subsection{Column Submatrix Labeling Rules}
\vspace{-2ex}
\renewcommand{\arraystretch}{1.5}
\begin{table}[H]
\caption{Column submatrix coloring rules to ensure equitable bicoloring among rows.\label{tab:bicoloring}}
\begin{tabular}
{|p{0.12\linewidth}|p{0.83\linewidth}|}
    \hline
     D.V.(s) & Labeling Rules\\ \hline
     $p$ & Verify there is no level (vertices with the same depth, ex. leaf set $L$) of $G_h$ missing. If so, relabel depth of selected vertices and repeat labeling procedure of $p$.\\ \hline
     $w$ & Determine the parity of the number of columns selected and assign alternating $red,~blue$ to columns according to their correspoding $v\in V$ depth. \\\hline
     $b$ & Repeat labeling procedure of $b$. \\\hline
     $s$ or $q$ & Use given column labels since we have identity matrices. \\\hline
     $p$ \& $w$ & Find the row with the most columns from the same set of $w$ with an invalid bicoloring (this row will only contain non-zeros in the same set). Switch the labels of half the non-zero columns of the chosen set. Repeat until no such row exists. \\\hline
     $p$ \& $b$ & Redo the labeling procedure for $p$ and $b$ independently. Columns of $p$ yield a bicoloring value in $\{0,1\}$, favoring $red$, and columns of $b$ yield a value in $\{0,1\}$, favoring $blue$.  \\\hline
     $w$ \& $s$ & Use given labels: each row has exactly one +1 and one -1 with the same color. \\\hline
     $b$ \& $q$ & Redo the labeling procedures of $b$ and $q$ starting with columns of $b$. \\\hline
     $p$ \& $s$/$q$;\newline$s$/$w$~\& $q$/$b$ & Use rules of each of the selected columns above as there exists no row such that columns of both decision variables contain non-zeros.\\\hline
\end{tabular}
\end{table}
\renewcommand{\arraystretch}{1}
As~\citet{ghoulia1962} stated every column submatrix must admit a valid equitable bicoloring for the matrix to be TU. In Table~\ref{tab:bicoloring} we enumerate the \emph{important} combinations of columns that may appear in a submatrix of $\baseq$ and provide an equitable coloring for such submatrices. For any submatrix containing columns of three or four different decision variables it is unnecessary to create a set of rules, rather we follow the rules of binary combinations. This follows from any row in the matrix containing non-zeros in columns of at most two different decision variables, as any constraint of $\baseq$ contains at most two decision variables.

\section{Computational Experiments}\label{compexpr}
Here we provide experiments on publicly available datasets to benchmark our four proposed models ($\mcfone$, $\mcftwo$, $\cutone$, $\cuttwo$) against five approaches from the literature: $\flowoct$ and its Benders' decomposition, $\bendersoct$ ($\milo$ approaches); and DL8.5, GOSDT+guesses, and QuantBnB (branch and bound approaches).

\subsection{Experimental Setup}
We run all experiments on an Intel(R) Core(TM) i7-9800X CPU (3.8Ghz, 19.25MB, 165W) using 1 core and 16GB RAM. Code is written in Python 3.9. $\milo$ formulations are solved using Gurobi 9.5 with a time limit of 3600 seconds; 600 seconds for branch and bound models. Datasets are publicly available at \url{http://archive.ics.uci.edu/ml/index.php} and code is available at \url{https://github.com/brandalston/OBCT}. We use 12 categorical and 4 numerical datasets from the UCI ML repository. As we assume the training dataset is in $\{0,1\}^{|F|}$ for our $\milo$ models we first perform an encoding process. For categorical datasets we use the standard one-hot encoding. For numerical datasets we perform a \emph{bucketization} procedure~\citep{lin2020,michini2020,verwer2019}. \emph{Bucketization} involves ordering the observations by any feature $f$, finding consecutive observations with different class labels (and different feature values), and defining a binary feature that has value 1 if and only if $x_f$ is less than the average of the two adjacent feature values.~\citet{lin2020} showed this procedure foregoes optimality.

\begin{figure}[H]
\centering
\begin{tikzpicture}[scale=0.25]
\begin{scope}[every node/.style={thick,draw}]
\node (0) at (-9,0) [circle] {$1$};
\node (1) at (0,0) [rectangle] {$q^i_a=0.75$};
\node (2) at (9,0) [rectangle, label={[align=center,xshift=-4mm, font=\fontsize{10.5}{0}\selectfont]above:I,}, label={[align=center,font=\fontsize{10.5}{0}\selectfont]above:II,}, label={[xshift=5mm, yshift=.75mm, align=center,font=\fontsize{10.5}{0}\selectfont]above:III}] {$q^i_b=0.25$};
\node (3) at (18,0) [rectangle, label={[align=center,yshift=.75mm,font=\fontsize{10.5}{0}\selectfont]above:I}] {$q^i_c=0.5$};
\node (4) at (27,0) [rectangle, label={[align=center,xshift=-3mm,font=\fontsize{10.5}{0}\selectfont]above:I,}, label={[align=center,xshift=3mm,font=\fontsize{10.5}{0}\selectfont]above:(III)}] [label={[align=center,font=\fontsize{11}{0}\selectfont]right:$s^i_v = 0.75$}] {$q^i_v=0.25$};
\end{scope}
\begin{scope}[>={Stealth[black]}]
\draw [->, dashed] (0) -- (1);
\draw [->] (1) -- (2);
\draw [->] (2) -- (3);
\draw [->] (3) -- (4);
\end{scope}
\end{tikzpicture}
\caption{Let $a, b, c$, and $v$ be nodes selected on the $1,v$-path of datapoint $i \in I$ at a fractional point in the branch and bound tree with $s^i_v$ and $q^i_u$ for $u \in P_v$ as defined. The 3 types of fractional separation cuts are indicated above (the III in parentheses is a a most violating cut considered but not added).\label{fig:fractypes}}
\end{figure}
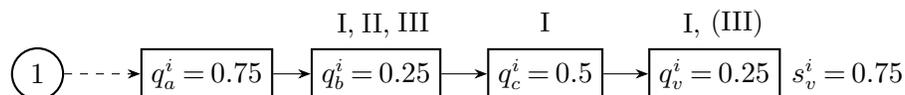

For each dataset we create 5 random 75-25\% train-test splits and train trees of depth $h \in \{2,3,4,5\}$. We do not consider the second objective function~\eqref{basemin}, minimizing the number of branching vertices, except for the case of the Pareto frontiers in Figure~\ref{fig:pareto}. No warm-starts unless otherwise stated. Models $\cutone$ and $\cuttwo$ introduce the on-the-fly cut constraints~\eqref{cut13} and~\eqref{cut23}, respectively, through three variations of user fractional cuts (outlined in Figure~\ref{fig:fractypes}), all up front, and at the root node of the branch and bound tree (Gurobi lazy parameter = 3). The best of the five variations is chosen. $\flowoct$ and $\bendersoct$ instances have a $\lambda$ value of zero. GOSDT+guesses uses an initial lower bound computed by the python 3.x package~\texttt{sklearn GradientBoosted} classifier. Quant-BnB uses an initial lower bound found by the process outlined in Section 3.2 of~\citet{mazumder2022}. Note Quant-BnB does not produce depth $h=\{4,5\}$ trees.

\begin{table}[H]
\renewcommand{\arraystretch}{1.2}
\caption{Dataset size ($|I|$), number of encoded features ($|F|$), number of classes ($|K|$) and featureset type: categorical or numerical ($\mathcal{C}/\mathcal{N}$).\label{tab:datasets}}
\fontsize{8.5}{9}\selectfont\centering
\begin{tabular}{c||cccccccccccccccc}\hline
     Dataset & soy & m3 & m1 & h.r. & m2 & h.v.84 & spect & b.c. & b.s. & t.t.t. & car & fico & iris & blood & climate & ion\\\hline
     $|I|$ & 47 & 122 & 124 & 132 & 169 & 232 & 267 & 277 & 625   & 958 & 1728 & 10459 & 150 & 747 & 540 & 351 \\
     $|F|$ & 45 & 15 & 15 & 15 & 15 & 16 & 22 & 38 & 20 & 27 & 20 & 19 & 43  & 122 & 1483 & 2260 \\
     $|K|$ & 4 & 2 & 2 & 3 & 2 & 2 & 2 & 2 & 3 & 2 & 4 & 2 & 3 & 2 & 2 & 2 \\
     Type & $\mathcal{C}$ & $\mathcal{C}$ & $\mathcal{C}$ & $\mathcal{C}$ & $\mathcal{C}$ & $\mathcal{C}$ & $\mathcal{C}$ & $\mathcal{C}$ & $\mathcal{C}$ & $\mathcal{C}$ & $\mathcal{C}$ & $\mathcal{C}$ & $\mathcal{N}$ & $\mathcal{N}$ & $\mathcal{N}$ & $\mathcal{N}$ \\\hline
\end{tabular}
\end{table}
\normalsize\renewcommand{\arraystretch}{1}

\subsection{Results}
\emph{Optimization Comparison}. Table~\ref{tab:timegap} summarizes in-sample optimization performance; solution time and optimality gap percentage. Our proposed formulations ($\mcfone,~\mcftwo,\cutone,~\cuttwo$) outperform the $\milo$ benchmark models ($\flowoct$ and $\bendersoct$) in 36 out of 64 test instances with an average speedup factor of $1.92 \pm 1.54$X. We define speedup factor as the ratio of solution time or gap. Comparing only flow-based $\milo$ models we see $\mcfone$ and $\mcftwo$ outperform $\flowoct$ in 47 out of 64 test instances with an average speedup factor of $2.89 \pm 2.39$X. Cut-based models $\cutone$ and $\cuttwo$ outperform $\flowoct$ in 52 out of 64 instances with a speed up factor of $2.91 \pm 2.33$X.~$\flowoct$ outperforms all of our proposed models in only 8 out of 64 test instances. Comparing against only $\bendersoct$ we see $\mcfone$ and $\mcftwo$ outperform in 32 out of 64 test instances with an average speedup factor of $2.93 \pm 5.59$X; $\cutone$ and $\cuttwo$ outperform in 33 out of 64 test instances with an average speedup factor of $3.57 \pm 7.00$X. $\cutone$ or $\cuttwo$ acheives the minimum solution time or gap in 27 out of 64 test instances; $\cuttwo$ achieving the minimum in 17 such instances, confirming theoretical results shown earlier. Notable significant improvements occur within numerical datasets. In 3 of the 16 test instances our models find an optimal solution within the time limit while $\flowoct$ and $\bendersoct$ report optimality gap $>5\%$. Further we report an improvement of solution time or gap in 13 of 16 test instances against $\flowoct$ and $\bendersoct$. Observe that for larger trees, ($h=4,5$), or larger datasets, ($|I| > 500$), we see that models $\cutone$ and $\cuttwo$ perform well highlighting the strength of cut constraints that are added on-the-fly; $\mcftwo$ remains competitive in such instances as it is a full model. Our flow-based models significantly outperform $\flowoct$, indicating the strength of their LO relaxations in practical applications. As expected, branch and bound formulations DL8.5, GOSDT+guesses and Quant-BnB are very fast; notable exceptions are the~\texttt{climate} and \texttt{ion} datasets where both DL8.5 and Quant-BnB were killed due to memory for depths $h=\{3,4,5\}$ and \texttt{ion}, $h=2$ where Quant-BnB found a solution in 484s, $\sim66$s faster than $\cuttwo$ (the only $\milo$ model to find a optimal solution within the 3600s time limit).

\begin{figure}[H]
\centering
    \includegraphics[width=.99\linewidth]{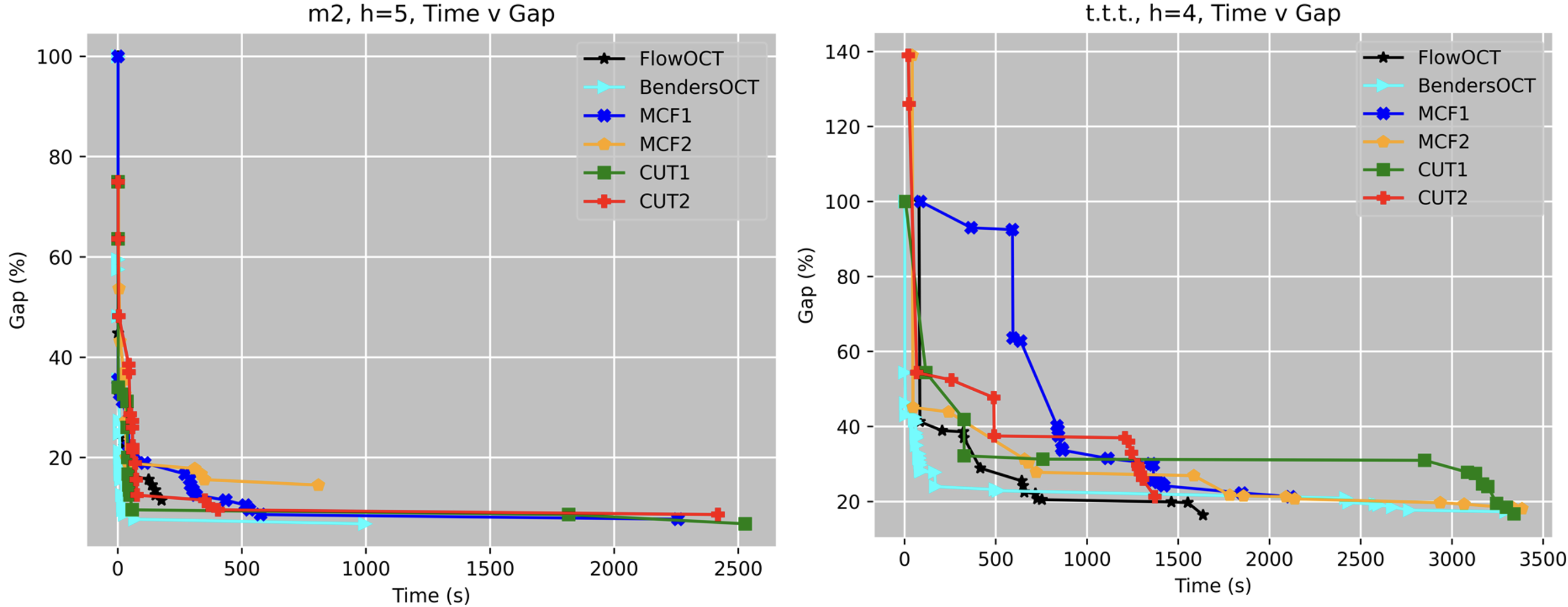}
    \caption{Reduction in optimality gap over time by each $\milo$ model for \texttt{m2} and \texttt{t.t.t.} datasets.\label{fig:instances}}
\end{figure}

Figure~\ref{fig:instances} summarizes reduction in in-sample optimality gap overtime. Models $\cutone$ and $\cuttwo$ introduce integral cut constraints that cut off the relaxation solution at the root node of the branch and bound tree. Observe models $\mcfone$ and $\mcftwo$ provide better incumbent solutions over $\cutone$ and $\cuttwo$, highlighting the strong LO relaxations of the full models, followed by the cut-models quickly reducing optimality gap, underscoring the effectiveness of our proposed cuts. It shows the strength of our cut-based formulations over flow-based formulations. Further, it confirms empirically the theoretical improvements in LO relaxations of our models shown in Section~\ref{theorycomp}, particularly with models $\mcftwo,~\cutone,$ and $\cuttwo$. Observe for the \texttt{m2} dataset (small test instance) all models reduce gap to within $\sim$5\% by~$\sim$1000s. For the \texttt{t.t.t.} dataset (large size instance) gap is reduced within $\sim$25\% by~$\sim$1500s, with $\mcftwo$ significantly outperforming in the beginning but stalling after $700s$, due to its strong LO relaxation but large size formulation.

\emph{Test Accuracy Comparison}. Table~\ref{tab:accuracy} summarizes the MILO models out-of-sample accuracy performance. Comparing our proposed models against only $\flowoct$ we see an improvement in 37 out of 64 test instances with an average increase of $2.00\%$. Comparing to $\bendersoct$ we see an improvement in 47 out of 64 test instances with an average increase of $2.05\%$. $\mcfone$ and $\mcftwo$ outperform $\flowoct$ and $\bendersoct$ in 29 out of 64 and 34 out of 64 test instances, respectively, with an average increase of $1.55\%$ and $1.80\%$, respectively. $\cutone$ and $\cuttwo$ outpeform $\flowoct$ and $\bendersoct$ in 35 out of 64 and 38 out of 64 test instances, respectively, with an average increase of $1.89\%$ and $2.14\%$, respectively. In any average comparison of the benchmark $\milo$ models to our proposed models in which the benchmarks win, we see our models stay within a $5\%$ decrease in accuracy. Comparing our proposed $\milo$ models to the branch and bound models we see an improvement in 34 out of 64 test instances with an average increase of $9.98\%$. $\mcfone$ and $\mcftwo$ out perform the branch and bound models in 31 out of 64 instances with an average increase of $9.97\%$, while $\cutone$ and $\cuttwo$ outperform in 34 out of 64 instances with an average increase of $9.82\%$. Notable improvements over branch and bound models occur when comparing our proposed models against GOSDT+guesses where we show a minimum improvement of $>15\%$ in 37 out of 64 test instances. Further, in only 8 instances do our proposed models have a $>5\%$ decrease in out-of-sample accuracy against all branch and bound benchmark models.

\begin{figure}[H]
\centering
    \includegraphics[width=.99\linewidth]{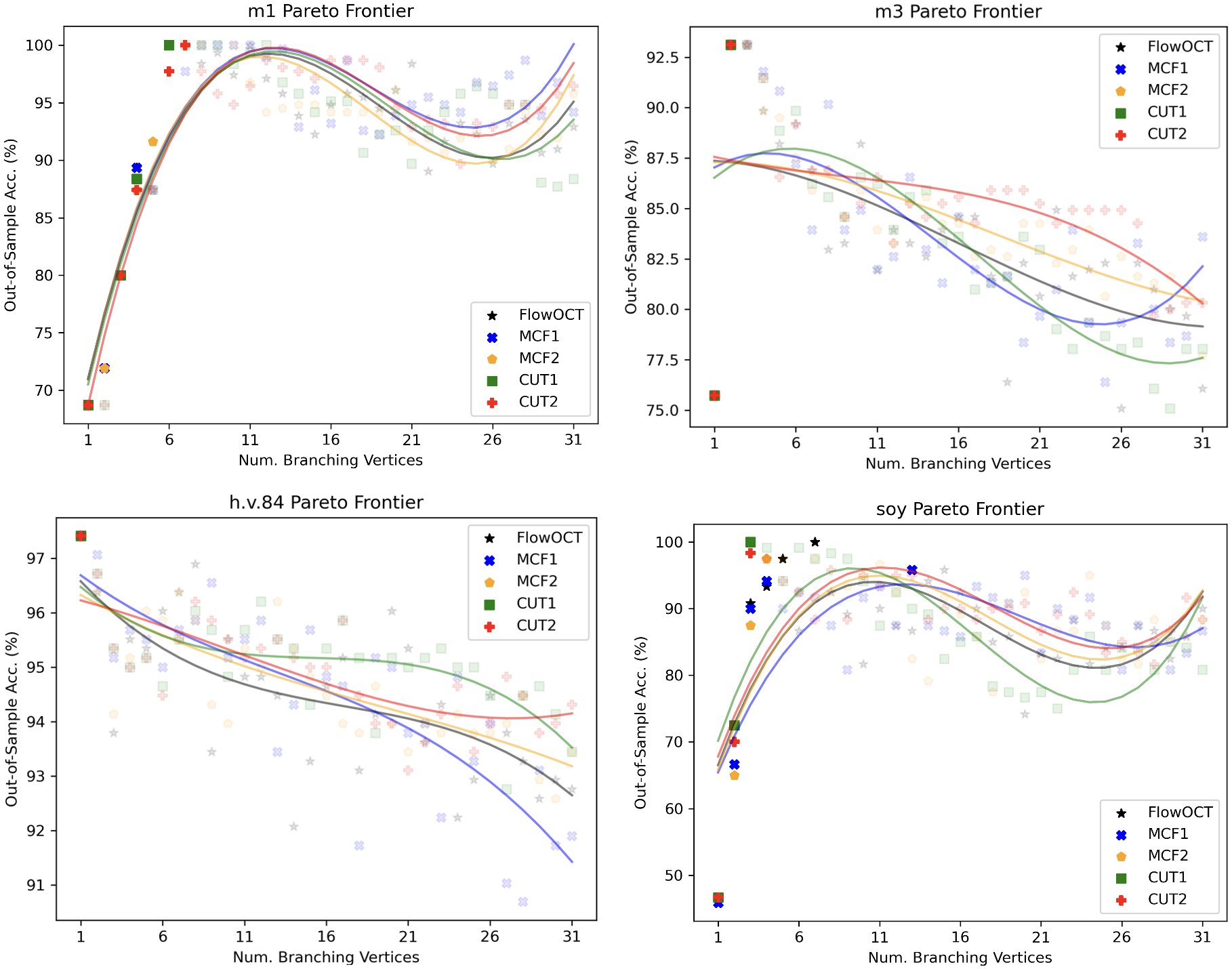}
    \caption{Pareto Frontiers for \texttt{m1, m3, h.v.84}, and \texttt{soy} datasets. Dominating points are in \textbf{bold}.}\label{fig:pareto}
\end{figure}

Figure~\ref{fig:pareto} summarizes our experiments of all $\milo$ formulations with objective function ~\eqref{basemax} and objective function~\eqref{basemin} used as an equality constraint on the number of branching vertices. Shown are out-of-sample accuracy (\%) average vs number of branching decisions and trend lines. Dominating and dominated Pareto points are illustrated in opaque and transparent colors, respectively. The solution for $k$ branching vertices is used as a warm-start for $k+1$ branching vertices for all models. Observe that $\mcftwo$, $\cutone$, $\cuttwo$ are the superior models when the maximum number of branching vertices is large. $\mcftwo$ generally performs well due to its strong LO relaxation and the fact the entire model is fed upfront, compared to lazy constraints for $\cutone$ and $\cuttwo$; these three models also perform well when the tree size increases, as expected. Additionally, one can see our models achieve higher out of sample accuracy compared to $\flowoct$ with the same or fewer branching vertices (particularly in \texttt{m3} and \texttt{h.v.84}). The Pareto frontiers confirm in $\milo$ formulations of optimal binary decision trees the observations of~\citet{murthy2004} that larger trees tend to perform inferior. Further they show fixing the topology of the tree does not necessarily forego optimality, highlighting the robustness and flexibility of $\milo$ models for decision trees.

\emph{Cut Constraints Comparison}.
Cut-based formulations $\cutone$ and $\cuttwo$ use a number of variations on implementing cut constraints~\eqref{cut13} and~\eqref{cut23}. We introduce integral cut constraints that cut off the relaxation solution at the root node and all integral cut constraints up front. Then we consider the cut constraints at fractional points in the branch and bound tree with three variations. The first type (I) adds all violating cuts for a datapoint in the $1,v$-path of a terminal vertex $v$; the second type (II) adds the first found violating cut in the $1,v$-path; the third type (III) adds the most violating cut, closest to the root of $G_h$, in the $1,v$-path. We consider a "heavy" set of user cuts (all violating cuts) and a "light" sets of user cuts (first found and most violating) due to~\citet{fischetti2017} who note adding too many fractional cuts may slow down solution time of $\milo$ formulations for the Steiner tree problem, which is highly related to decision trees.

Tables~\ref{tab:cut1table} and~\ref{tab:cut2table} compare our variations for implementing cut constraints on a subset of test datasets. The UF column contains the solution time or gap of the CUT model with all integral cut constraints introduced upfront. The other columns represent solution time or gap ratios to the UF column. Decreases (increases) in solution time or gap are indicated in \textbf{bold} (\textcolor[HTML]{757171}{light}). The LAZY column represents integral constraints introduced to cut off the relaxation solution at the root node of the branch and bound tree. Both tables highlight adding our cut constraints on-the-fly does help improve solution time over adding all cut constraints upfront; a speedup factor of up to $\sim$3X for even large size instances. Tables~\ref{tab:cut1table} and~\ref{tab:cut2table} confirm the observations of~\citet{fischetti2017} for optimal binary decision tree $\milo$ formulations as instances are reported with a $\sim$92X slowdown when our fractional cuts are applied! While there are improvements using on-the-fly cuts for large trees, our testing does not suggest a convention on which procedure is most effective. Note that cuts obtained by solving the fractional separation problem are most effective for smaller trees. Results shown only consider fractional cuts at the root node of the branch and bound tree as our testing showed marginal improvement when user fractional cuts were added at all nodes.

\begin{table}[H]
\caption{Average solution time, optimality gap \%-age (in parentheses) if TL reached. Best in \textbf{bold}.\label{tab:timegap}}
\fontsize{8.5}{9}\selectfont\centering
\renewcommand{\arraystretch}{1.05}
\begin{tabular}{l||ccccccccc}
Dataset & $\flowoct$ & $\bendersoct$ & $\mcfone$ & $\mcftwo$ & $\cutone$ & $\cuttwo$ & DL8.5 & GOSDT+g & Q-BnB \\\hline

& \multicolumn{9}{c}{$h$=2}\\\hline
soy & 0.08 & 0.20 & 0.10 & 0.06 & \textbf{0.04} & 0.07 & $<0.00$  & 0.05 & 0.63\\
m3 & 0.51 & \textbf{0.23} & 0.94 & 1.02 & 0.56 & 0.56 & $<0.00$  & 0.01 & 0.01   \\
m1& 0.74   & \textbf{0.48} & 1.68 & 1.12 & 0.67 & 0.73 & $<0.00$  & 0.01 & 0.01   \\
h.r.  & 0.69 & 0.83 & 2.69 & 1.01 & \textbf{0.62} & \textbf{0.62} & $<0.00$ & 0.04 & 0.01\\
m2   & 6.76   & 3.12 & 5.83 & 3.40   & \textbf{1.87}   & 1.92 & $<0.00$  & 0.01 & 0.01   \\
h.v.84 & 0.76   & \textbf{0.48} & 2.97 & 1.97   & 1.06   & 1.07 & $<0.00$  & 0.01 & 0.75   \\
spect   & 9.97   & 4.14 & 15.90   & 13.96  & 2.84   & \textbf{2.63} & $<0.00$  & 0.01 & 0.10   \\
b.c. & 23.69  & 17.19   & 45.94 & 7.79 & 7.68 & \textbf{7.43} & $<0.00$  & 0.01 & 0.11   \\
b.s. & 10.51  & 6.60 & 21.68   & 17.16 & 7.29   & \textbf{7.23} & $<0.00$  & 0.06 & 0.04   \\
t.t.t.  & 490.04 & 139.20  & 327.18  & 96.10  & \textbf{93.31}  & 104.08  & 0.01  & 0.02 & 0.68   \\
car  & 83.46  & \textbf{32.08}   & 111.03  & 45.28  & 38.99  & 35.97   & $<0.00$  & 0.12 & 0.08   \\
fico & 944.53 & 2414.30 & (25.05) & \textbf{858.20} & 955.18 & 1408.68 & 0.01  & 0.10 & 0.30   \\
iris & 0.43   & 0.33 & 1.22 & 1.02   & 0.41   & \textbf{0.39} & $<0.00$  & 0.04 & 0.83   \\
blood   & 427.64 & 1638.88 & 399.62  & 632.70 & 246.76 & \textbf{56.89}   & 0.05  & 0.04 & 2.14   \\
climate & (6.47) & (6.30)   & (5.10)   & (4.15) & (4.25) & \textbf{935.43}  & 9.68  & 0.03 & 211.63 \\
ion & (9.40)  & (9.49)  & (5.41)  & (6.25) & (4.25) & \textbf{550.12}  & 16.48 & 0.04 & 484.58 \\ \hline

& \multicolumn{9}{c}{$h$=3}\\\hline
soy  & 0.16 & 0.20 & \textbf{0.08} & 0.09 & \textbf{0.08} & \textbf{0.08} & 0.01 & 0.05 & ERR   \\
m3   & 91.98   & \textbf{15.70}   & 105.15  & 92.57   & 66.25   & 91.99   & $<0.00$ & 0.01 & 0.06  \\
m1   & 32.65   & \textbf{4.33} & 35.65   & 35.19   & 25.44   & 25.73   & $<0.00$ & 0.01 & 0.80  \\
h.r.  & 13.86   & \textbf{7.86} & 59.02   & 31.50   & 40.76   & 26.83   & 0.01 & 0.05 & 0.08  \\
m2   & 2935.01 & 1205.76 & 1898.05 & \textbf{469.60}  & 1014.42 & 804.51  & 0.01 & 0.02 & 0.08  \\
h.v.84 & 200.91  & \textbf{80.30}   & 167.30  & 127.05  & 173.64  & 133.14  & 0.01 & 0.01 & 4.71  \\
spect   & 2790.05 & 2162.11 & 1760.49 & 1736.55 & \textbf{942.73}  & 1385.65 & 0.02 & 0.02 & 2.50  \\
b.c. & (73.58) & \textbf{(12.07)} & (17.99) & (13.22) & (12.98) & (13.91) & 0.09 & 0.02 & 1.39  \\
b.s. & 1863.60 & \textbf{895.25}  & 1651.92 & 1384.10 & 1235.88 & 1232.90 & 0.02 & 0.07 & 0.42  \\
t.t.t.  & $(>100)$   & \textbf{(18.57)} & (30.84) & (29.01) & (28.45) & (29.36) & 0.07 & 0.03 & 1.22  \\
car  & (78.89) & \textbf{(5.07)}  & (23.29) & (18.32) & (18.26) & (17.5)  & 0.03 & 0.17 & 0.69  \\
fico & (98.31) & (39.95) & \textbf{(39.51)}  & \textbf{(39.51)} & (41.47) & (41.43) & 0.07 & 0.14 & 2.03  \\
iris & 15.55   & 117.64  & \textbf{3.63} & 6.73 & 4.29 & 3.67 & 0.04 & 0.05 & 1.99  \\
blood & (21.46) & (23.68) & (19.52) & (18.87) & (18.31) & \textbf{(14.14)} & 5.11 & 0.06 & 85.65 \\
climate & (5.10) & (4.88)  & (5.43)  & (5.04)  & (18.31) & \textbf{(4.49)}  & MEM  & 0.72 & MEM   \\
ion & (5.89) & (7.90) & (6.31) & (5.97) & \textbf{(4.60)} & (6.14)  & MEM  & 0.62 & MEM \\ \hline

 & \multicolumn{9}{c}{$h$=4}\\\hline
soy  & 0.33 & 0.36 & 0.28 & 0.27 & \textbf{0.22} & \textbf{0.22} & $<0.00$   & 0.05   & N/A \\
m3   & 2210.76 & 2381.33 & 2186.47 & 2197.56 & \textbf{1733.84} & 1924.94 & 0.02   & 0.02   & N/A \\
m1   & 15.89   & \textbf{2.13} & 22.63   & 8.92 & 9.87 & 13.01   & $<0.00$   & 0.01   & N/A \\
h.r. & 2430.02 & 2422.33 & 1819.18 & 1415.03 & 1512.81 & \textbf{1270.11} & 0.03 & 0.07   & N/A \\
m2   & (17.23) & \textbf{(12.95)} & (18.36) & (17.43) & (18.05) & (18.60)  & 0.06   & 0.02   & N/A \\
h.v.84 & 2882.22 & 2882.02 & 2169.07 & \textbf{1087.32} & 1144.08 & 1209.22 & 0.02   & 0.02   & N/A \\
spect   & (35.36) & \textbf{(4.21)}  & (4.70)   & (4.48)  & (4.59)  & (4.36)  & 0.15   & 0.02   & N/A \\
b.c. & (16.05) & (15.67) & (17.11) & (15.27) & (15.52) & \textbf{(15.02)} & 1.36 & 0.03 & N/A \\
b.s.   & (97.26) & \textbf{(9.88)}  & (21.67) & (20.67) & (18.36) & (19.26) & 0.20   & 0.10   & N/A \\
t.t.t. & $(>100)$ & \textbf{(19.75)} & (22.21) & (19.52) & (19.20)  & (19.75) & 0.96 & 0.04 & N/A \\
car  & $(>100)$   & \textbf{(20.65)} & (28.27) & (21.44) & (22.73) & (21.90)  & 0.30   & 0.22   & N/A \\
fico & (99.57) & \textbf{(41.48)} & $(>100)$ & (60.95) & $(>100)$ & $(>100)$ & 0.69 & 0.17 & N/A \\
iris & 7.01 & \textbf{1.73} & 18.47   & 13.03   & 10.81   & 19.07   & 0.10   & 0.06   & N/A \\
blood   & (21.82) & (23.82) & \textbf{(21.07)} & (21.20)  & (21.36) & (21.71) & 292.29 & 34.46  & N/A \\
climate & (5.06)  & (5.70)   & (3.65)  & \textbf{(3.43)}  & (21.93) & (5.23)  & MEM & 0.62   & N/A \\
ion & (4.46)  & (7.83)  & (4.71)  & (3.47)  & \textbf{(3.42)}  & \textbf{(3.42)}  & MEM & 406.39 & N/A \\\hline

 & \multicolumn{9}{c}{$h$=5}\\ \hline
soy  & 0.65 & \textbf{0.49} & 0.52 & 0.65 & 0.49 & 0.50 & $<0.00$   & 0.05   & N/A \\
m3   & 187.22  & \textbf{42.62}   & 219.00  & 142.03  & 79.16   & 136.59  & 0.01   & 0.02   & N/A \\
m1   & 8.46 & \textbf{1.51} & 16.60   & 13.78   & 12.06   & 12.52   & $<0.00$   & 0.01   & N/A \\
h.r. & (8.86) & (9.06) & (9.53) & \textbf{(8.66)} & (9.12) & (9.82) & 0.14 & 0.09 & N/A \\
m2   & (11.71) & (10.38) & (9.20)   & (8.48)  & (9.20)   & \textbf{(8.45)}  & 0.15   & 0.03   & N/A \\
h.v.84 & 471.55  & 555.25  & \textbf{62.14} & 76.84 & 115.90  & 116.20  & $<0.00$ & 0.02   & N/A \\
spect  & (55.49) & (4.87)  & (4.61)  & (5.85)  & (5.30)   & \textbf{(4.17)}  & 0.72   & 0.03   & N/A \\
b.c. & (11.68) & (11.34) & (10.37) & (11.69) & (10.15) & \textbf{(9.90)} & 12.74  & 0.04 & N/A \\
b.s. & $(>100)$ & (21.63) & (21.44) & \textbf{(21.12)} & (21.32) & (21.56) & 1.31   & 0.14   & N/A \\
t.t.t.  & $(>100)$ & (14.24) & (13.04) & \textbf{(13.00)} & (16.27) & (14.81) & 7.02 & 0.06 & N/A \\
car  & (83.71) & \textbf{(18.03)} & (40.15) & (32.01) & (39.44) & (34.43) & 1.83   & 0.28   & N/A \\
fico & (40.71) & \textbf{(40.64)} & $(>100)$ & $(>100)$ & $(>100)$ & $(>100)$ & 4.46 & 0.20 & N/A \\
iris & 10.01   & \textbf{1.93} & 32.87   & 22.26   & 29.91   & 31.96   & 0.08   & 0.06   & N/A \\
blood   & (19.84) & (22.35) & (20.42) & (20.20)  & \textbf{(19.52)} & (20.97) & TL & 490.50 & N/A \\
climate & (6.65)  & \textbf{(4.05)}  & (4.17)  & (4.25)  & (20.04) & (5.62)  & MEM & 0.99   & N/A \\
ion & (5.43)  & (8.92)  & (2.43)  & \textbf{3509.67} & (4.17)  & (5.13)  & MEM & MEM & N/A \\\hline
\end{tabular}
\end{table}
\normalsize

\begin{table}[H]
\caption{Average out-of-sample accuracy (\%). Best in \textbf{bold}.\label{tab:accuracy}}
\fontsize{8.5}{9}\selectfont\centering
\renewcommand{\arraystretch}{1.05}
\begin{tabular}{l||ccccccccc}
    Dataset & $\flowoct$ & $\bendersoct$ & $\mcfone$ & $\mcftwo$ & $\cutone$ & $\cuttwo$ & DL8.5 & GOSDT+g & Q-BnB \\\hline
    
    & \multicolumn{9}{c}{$h$=2}\\\hline
    soy & \textbf{100} & 98.33 & 98.33 & \textbf{100} & \textbf{100} & \textbf{100} & 96.67 & 18.33 & 91.67 \\
    m3 & \textbf{94.19} & \textbf{94.19} & \textbf{94.19} & \textbf{94.19} & \textbf{94.19} & \textbf{94.19} & \textbf{94.19} & 47.74 & 93.55 \\
    m1 & 74.84 & 74.84 & 74.84 & 74.84 & 74.84 & 74.84 & \textbf{82.58} & 49.68 & 74.84 \\
    h.r. & 44.24 & 44.24 & 44.24 & 44.24 & 44.24 & 44.24 & 48.48 & 25.45 & \textbf{50.91} \\
    m2 & 54.42 & \textbf{56.28} & 55.81 & 55.81 & 55.81 & \textbf{56.28} & 54.42 & 55.35 & 54.88 \\
    h.v.84 & \textbf{95.52} & \textbf{95.52} & \textbf{95.52} & \textbf{95.52} & \textbf{95.52} & \textbf{95.52} & \textbf{95.52} & 62.41 & 95.23 \\
    spect & 77.01 & 77.01 & 77.01 & 77.01 & 77.01 & 77.01 & 80.30 & \textbf{83.28} & 74.63 \\
    b.c. & \textbf{74.00} & 72.00 & 72.00 & 73.14 & 72.57 & 72.57 & 69.71 & 67.71 & 68.00 \\
    b.s. & 64.20 & 64.20 & 64.20 & 64.20 & 64.20  & 64.20  & 65.61 & 43.18 & \textbf{65.73} \\
    t.t.t. & \textbf{68.58} & 68.33 & \textbf{68.58} & \textbf{68.58} & \textbf{68.58}  & \textbf{68.58} & 67.08 & 66.50 & 65.67 \\
    car & \textbf{77.78} & \textbf{77.78} & \textbf{77.78} & \textbf{77.78} & \textbf{77.78} & \textbf{77.78} & \textbf{77.78} & 68.98 & 77.50 \\
    fico & 70.73 & 70.73 & \textbf{70.80} & 70.73 & 70.73 & 70.73 & 70.73 & 52.93 & 70.45 \\
    iris  & \textbf{94.21} & \textbf{94.21} & \textbf{94.21} & \textbf{94.21} & \textbf{94.21} & \textbf{94.21} & \textbf{94.21} & 13.16 & 93.16 \\
    blood & 75.61 & 76.26 & 76.26 & 76.36 & 76.36 & 76.26 & 76.36 & \textbf{77.01} & 75.08 \\
    climate & 90.81 & 89.48 & 89.78 & 90.22 & 90.22 & 90.52 & \textbf{93.04} & 92.44 & 90.81 \\
    ion & 90.91 & 91.36 & 91.36 & 90.91 & 91.10  & \textbf{91.59} & 83.86 & 27.50 & 85.45 \\ \hline

    & \multicolumn{9}{c}{$h$=3}\\\hline
    soy & \textbf{98.33} & 95.00 & \textbf{98.33} & 96.67 & \textbf{98.33} & \textbf{98.33} & 96.67 & 30.00 & ERR \\
    m3 & \textbf{92.90} & 92.26 & 92.26 & 92.26 & \textbf{92.90} & \textbf{92.90}& 91.61 & 56.13 & 91.61 \\
    m1 & 84.52 & \textbf{87.74} & 87.10  & 85.81 & 86.45 & 87.10  & 87.10 & 53.55 & 83.23 \\
    h.r. & 56.97 & 56.36 & 56.36 & 56.36 & 56.36 & 56.36 & 59.39 & 40.61 & \textbf{62.42} \\
    m2 & 68.84 & 66.05 & 64.65 & 67.44 & 69.3  & \textbf{69.77} & 68.84 & 58.60 & 66.51 \\
    h.v.84 & 93.10  & 92.41 & 92.76 & 92.76 & 93.10  & 92.76 & \textbf{94.14} & 43.10 & 93.94 \\
    spect & 77.31 & 77.31 & 77.61 & 77.61 & 77.61 & 77.61 & 76.72 & \textbf{83.28} & 69.55 \\
    b.c. & 71.43 & 67.43 & 69.71 & 71.71 & 69.71 & \textbf{72.00} & 65.43 & 67.71 & 67.71 \\
    b.s. & 67.64 & 67.64 & 67.26 & 67.64 & 67.64 & 67.64 & 70.45 & 43.18 & \textbf{71.34} \\
    t.t.t. & 73.17 & 73.75 & 74.08 & \textbf{74.33} & 74.08 & 74.33 & 72.75 & 66.50 & 70.25 \\
    car & 78.43 & 78.52 & 78.19 & 78.52 & 78.94 & 78.94 & 79.31 & 68.98 & \textbf{79.72} \\
    fico & 70.97 & 70.65 & 70.88 & 70.97 & 69.76 & 69.77 & \textbf{71.14} & 52.93 & 70.96 \\
    iris  & 94.21 & 95.26 & 94.74 & 94.21 & 95.26 & 94.74 & \textbf{97.37} & 17.89 & 94.08 \\
    blood & 77.22 & 77.54 & 76.2  & 76.04 & 77.65 & 77.43 & 78.61 & 77.01 & \textbf{79.36} \\
    climate & 89.33 & 89.63 & 88.15 & 89.78 & \textbf{90.37} & 89.93 & MEM & 68.44 & MEM \\
    ion & 86.14 & 88.64 & 86.59 & 86.14 & \textbf{89.55} & 86.14 & MEM  & 31.14 & MEM \\ \hline

    & \multicolumn{9}{c}{$h$=4}\\\hline
    soy & \textbf{98.33} & 95.00 & 96.67 & \textbf{98.33} & \textbf{98.33} & 96.67 & 85.00 & 30.00 & N/A \\
    m3 & \textbf{92.26} & 90.97 & 89.03 & 89.68 & 90.32 & \textbf{92.26} & 89.03 & 57.42 & N/A \\
    m1 & 98.06 & \textbf{100} & \textbf{100} & \textbf{100} & \textbf{100} & \textbf{100} &\textbf{100} & 52.26 & N/A \\
    h.r. & 67.27 & 65.45 & 64.85 & 66.06 & 67.27 & 67.27 & \textbf{70.30} & 30.30 & N/A \\
    m2 & 56.74 & \textbf{66.05} & 62.33 & 61.86 & 65.58 & 65.12 & 60.93 & 62.79 & N/A \\
    h.v.84 & 95.86 & 95.17 & 94.83 & 96.21 & \textbf{96.55} & \textbf{96.55} & 94.48 & 73.10 & N/A \\
    spect & 80.30  & 79.40  & 77.61 & 81.19 & 79.7  & 80.00 & 74.33 & \textbf{83.28} & N/A \\
    b.c. & 68.57 & 68.86 & 69.14 & 69.43 & 70.71 & \textbf{71.43} & \textbf{71.43} & 67.71 & N/A \\
    b.s. & 70.57 & 69.30  & 70.83 & 71.72 & \textbf{72.48} & 71.46 & 71.34 & 43.18 & N/A \\
    t.t.t. & 80.42 & 81.33 & 76.17 & \textbf{80.92} & 80.50  & 80.50  & 81.50 & 66.50 & N/A \\
    car & \textbf{81.85} & 80.79 & 78.15 & 80.51 & 80.28 & 80.23 & 82.41 & 68.98 & N/A \\
    fico & 70.20  & 70.11 & 53.06 & 61.93 & 53.06 & 53.06 & \textbf{71.27} & 52.93 & N/A \\
    iris  & 93.68 & 94.74 & 93.68 & 92.63 & \textbf{96.32} & 95.79 & 94.21 & 14.74 & N/A \\
    blood & 77.22 & 77.75 & 77.01 & 75.08 & 76.04 & \textbf{78.07} & 75.94 & 59.79 & N/A \\
    climate & 89.33 & 88.44 & 90.37 & \textbf{91.41} & 89.93 & 90.81 & MEM  & 82.81 & N/A \\
   ion & 87.05 & 85.68 & 86.14 & 86.82 & \textbf{89.55} & 89.09 & MEM  & 31.44 & N/A \\ \hline

    & \multicolumn{9}{c}{$h$=5}\\ \hline
    soy & 98.33 & 90.00 & \textbf{100}   & 96.67 & 96.67 & \textbf{100}   & 86.67 & 30.00 & N/A \\
    m3 & 88.39 & 87.10  & 89.03 & 89.68 & \textbf{91.61} & 87.74 & 82.58 & 54.19 & N/A \\
    m1 & \textbf{100}   & \textbf{100}   & 98.71 & \textbf{100}   & \textbf{100}   & \textbf{100}   & 92.26 & 49.03 & N/A \\
    h.r. & 70.91 & 68.48 & 64.24 & 70.91 & \textbf{75.76} & \textbf{75.76} & 66.06 & 29.09 & N/A \\
    m2 & 58.14 & 59.07 & \textbf{65.58} & \textbf{65.58} & 64.65 & 64.65 & 62.33 & 52.09 & N/A \\
    h.v.84 & 93.79 & 93.79 & 93.79 & 95.17 & \textbf{95.86} & 95.52 & 94.83 & 47.59 & N/A \\
    spect & 81.19 & 77.91 & 77.31 & 80.00 & 79.40  & 80.00 & 74.03 & \textbf{83.28} & N/A \\
    b.c. & 70.86 & 69.43 & 70.00 & 69.43 & \textbf{71.43} & \textbf{71.43} & 68.29 & 67.71 & N/A \\
    b.s. & 73.12 & 69.30  & 72.48 & 71.34 & \textbf{73.25} & 73.25 & 72.23 & 43.18 & N/A \\
    t.t.t. & 81.17 & 84.75 & 82.00 & \textbf{84.92} & 81.75 & 84.25 & 87.00 & 66.50 & N/A \\
    car & 83.24 & 82.41 & 72.08 & 75.60  & 71.25 & 75.00 & \textbf{86.99} & 68.98 & N/A \\
    fico & 68.82 & 70.02 & 53.06 & 61.93 & 53.06 & 53.06 & \textbf{71.21} & 52.93 & N/A \\
    iris  & 93.16 & 94.74 & \textbf{95.79} & 93.16 & 95.26 & 93.68 & 94.21 & 16.32 & N/A \\
    blood & 76.04 & 77.22 & 76.36 & 75.19 & \textbf{78.40}  & 77.22 & 75.94 & 46.31 & N/A \\
    climate & 88.89 & 89.63 & \textbf{90.37} & 87.70  & 89.48 & 89.33 & MEM  & 92.44 & N/A \\
    ion & 85.00 & 86.36 & \textbf{88.18} & 86.14 & 84.32 & 85.45 & MEM  & MEM  & N/A \\\hline
\end{tabular}
\end{table}
\normalsize

\vspace{-6ex}
\begin{table}[H]
\parbox{.49\linewidth}{
\fontsize{8.5}{9}\selectfont
\caption{$\cutone$ fractional separation ratios.\label{tab:cut1table}}
\begin{tabular}{|l||c||c||cccc|}\hline
    Dataset & $G_h$ & UF & LAZY & \Romannum{1} & \Romannum{2} & \Romannum{3} \\ \hline
    
    \multirow{4}{*}{m3} & 2 & 1.52 & \textbf{0.37} & \textbf{0.38} & \textbf{0.38} & \textbf{0.38} \\ 
    & 3 & 66.25 & \textcolor[HTML]{757171}{1.22} & \textcolor[HTML]{757171}{1.13} & \textcolor[HTML]{757171}{1.86} & \textcolor[HTML]{757171}{1.85} \\ 
    & 4 & 2222.62 & \textbf{0.78} & \textbf{0.91} & \textbf{0.89} & \textbf{0.89} \\ 
    & 5 & 253.98 & \textbf{0.48} & \textbf{0.31} & \textbf{0.50} & \textbf{0.51} \\ \hline

    \multirow{4}{*}{h.v.84} & 2 & 2.89 & \textbf{0.37} & \textbf{0.37} & \textbf{0.37} & \textbf{0.37} \\ 
    & 3 & 173.64 & \textcolor[HTML]{757171}{1.46} & \textcolor[HTML]{757171}{1.54} & \textcolor[HTML]{757171}{1.67} & \textcolor[HTML]{757171}{1.66} \\ 
    & 4 & 1625.82 & \textcolor[HTML]{757171}{1.07} & \textbf{0.70} & \textbf{0.80} & \textbf{0.80} \\ 
    & 5 & 115.90 & \textcolor[HTML]{757171}{1.37} & \textcolor[HTML]{757171}{1.22} & \textcolor[HTML]{757171}{1.07} & \textcolor[HTML]{757171}{1.07} \\ \hline

    \multirow{4}{*}{b.c.} & 2 & 18.72 & \textbf{0.75} & \textbf{0.41} & \textbf{0.41} & \textbf{0.41} \\ 
    & 3 & (12.98) & \textcolor[HTML]{757171}{1.22}	& \textcolor[HTML]{757171}{1.17} & \textcolor[HTML]{757171}{1.23} & \textcolor[HTML]{757171}{1.23} \\ 
    & 4 & (15.52) & \textcolor[HTML]{757171}{1.03} & \textcolor[HTML]{757171}{1.07} & \textcolor[HTML]{757171}{1.01} & \textcolor[HTML]{757171}{1.01} \\ 
    & 5	& (10.98) & \textbf{0.92} & \textcolor[HTML]{757171}{1.04} & \textcolor[HTML]{757171}{1.01} & \textcolor[HTML]{757171}{1.01} \\\hline

    \multirow{4}{*}{fico} & 2 & 955.18 & \textcolor[HTML]{757171}{3.77} & \textcolor[HTML]{757171}{3.58} & \textcolor[HTML]{757171}{3.73} & \textcolor[HTML]{757171}{3.71} \\ 
    & 3 & (41.47) & \textcolor[HTML]{757171}{5.30} & \textcolor[HTML]{757171}{6.70} & \textcolor[HTML]{757171}{7.35} & \textcolor[HTML]{757171}{7.35} \\ 
    & 4 & (106.89) & \textcolor[HTML]{757171}{3.68} & \textcolor[HTML]{757171}{3.69} & \textcolor[HTML]{757171}{3.68} & \textcolor[HTML]{757171}{3.68} \\ 
    & 5 & (106.94) & \textcolor[HTML]{757171}{60.45} & \textcolor[HTML]{757171}{91.69} & \textcolor[HTML]{757171}{52.28} & \textcolor[HTML]{757171}{35.67} \\ \hline

    \multirow{4}{*}{iris} & 2  & 1.32  & \textbf{0.39}  & \textbf{0.31}  & \textbf{0.31}  & \textbf{0.31} \\
    & 3 & 8.16  & \textbf{0.53}  & \textbf{0.67}  & \textbf{0.66}  & \textbf{0.67} \\
    & 4 & 27.73 & \textbf{0.39}  & \textbf{0.67}  & \textbf{0.68}  & \textbf{0.69} \\
    & 5 & 30.34 & \textcolor[HTML]{757171}{1.19}  & \textbf{0.99}  & 1.00  & 1.00 \\ \hline
    
    \multirow{4}{*}{blood} & 2  & 246.76 &	\textcolor[HTML]{757171}{1.82} &	\textcolor[HTML]{757171}{1.84} & \textcolor[HTML]{757171}{1.89} & \textcolor[HTML]{757171}{1.87} \\
    & 3 & (18.31) & \textcolor[HTML]{757171}{1.14}  & \textcolor[HTML]{757171}{1.09}  & \textcolor[HTML]{757171}{1.08}  & \textcolor[HTML]{757171}{1.09} \\
    & 4 & (21.93)  & \textcolor[HTML]{757171}{1.01}  & \textbf{0.97}  & \textbf{0.97}  & \textbf{0.97} \\
    & 5 & (20.04)  & \textcolor[HTML]{757171}{1.02}  & \textbf{0.97}  & \textbf{0.97}  & \textbf{0.97} \\ \hline
\end{tabular}}
\parbox{.49\linewidth}{
    \fontsize{8.5}{9}\selectfont
    \caption{$\cuttwo$ fractional separation ratios.\label{tab:cut2table}}
    \begin{tabular}[H]{|l||c||c||cccc|} \hline
    Dataset & $G_h$ & UF & LAZY & \Romannum{1} & \Romannum{2} & \Romannum{3} \\\hline 
    
    \multirow{4}{*}{m3} & 2 & 1.28 & \textbf{0.44} & \textbf{0.46} & \textbf{0.45} & \textbf{0.46} \\ 
    & 3 & 114.73 & \textbf{0.80} & \textbf{0.92} & \textbf{0.97} & \textbf{0.97} \\ 
    & 4 &  2186.42 & \textbf{0.88} & \textcolor[HTML]{757171}{1.02} & \textbf{0.90} & \textbf{0.90} \\ 
    & 5 & 383.06 & \textbf{0.36} & \textbf{0.44} & \textbf{0.36} & \textbf{0.37} \\ \hline
    
    \multirow{4}{*}{h.v.84} & 2 & 2.65 & \textbf{0.40} & \textbf{0.41}	& \textbf{0.40} & \textbf{0.41} \\ 
    & 3 & 133.14 & \textcolor[HTML]{757171}{1.97} & \textcolor[HTML]{757171}{1.78} & \textcolor[HTML]{757171}{1.74} & \textcolor[HTML]{757171}{1.75} \\ 
    & 4 & 1861.33 & \textbf{0.93} & \textbf{0.65} & \textbf{0.71} & \textbf{0.71} \\ 
    & 5 & 116.20 & \textcolor[HTML]{757171}{1.39} & \textcolor[HTML]{757171}{1.76} & \textcolor[HTML]{757171}{1.95} & \textcolor[HTML]{757171}{1.96} \\ \hline 
    
    \multirow{4}{*}{b.c.} & 2 & 65.27 & \textbf{0.22} & \textbf{0.11} & \textbf{0.11} & \textbf{0.11} \\ 
    & 3 & (15.74) & \textcolor[HTML]{757171}{1.07} & \textbf{0.89} & \textbf{0.88} & \textbf{0.88} \\ 
    & 4 & (15.02) &	\textcolor[HTML]{757171}{1.08} & \textcolor[HTML]{757171}{1.09} & \textcolor[HTML]{757171}{1.09} &	\textcolor[HTML]{757171}{1.08} \\ 
    & 5 & (10.66) & \textbf{0.93} & \textcolor[HTML]{757171}{1.06} & \textbf{0.99} & \textbf{0.94} \\ \hline
 
    \multirow{4}{*}{fico} & 2 & 1408.68 & \textcolor[HTML]{757171}{2.41} & \textcolor[HTML]{757171}{2.40} & \textcolor[HTML]{757171}{2.41} & \textcolor[HTML]{757171}{2.40} \\ 
    & 3 & (41.43) & \textcolor[HTML]{757171}{4.10} & \textcolor[HTML]{757171}{3.35} & \textcolor[HTML]{757171}{4.93} & \textcolor[HTML]{757171}{4.93} \\ 
    & 4 & (106.73) & \textcolor[HTML]{757171}{3.46} & \textcolor[HTML]{757171}{3.36} & \textcolor[HTML]{757171}{3.62} & \textcolor[HTML]{757171}{3.49} \\ 
    & 5 & (106.86) & \textcolor[HTML]{757171}{5.34} & \textcolor[HTML]{757171}{21.95} & \textcolor[HTML]{757171}{21.95} & \textcolor[HTML]{757171}{19.98} \\ \hline

    \multirow{4}{*}{iris} & 2 & 1.29 & \textbf{0.31}  & \textbf{0.30}  & \textbf{0.30}  & \textbf{0.30} \\
    & 3 & 7.98 & \textbf{0.46}  & \textbf{0.50}  & \textbf{0.49}  & \textbf{0.50} \\
    & 4 & 32.72 & \textbf{0.58}  & \textbf{0.71}  & \textbf{0.71}  & \textbf{0.70} \\
    & 5 & 50.23 & \textbf{0.64}  & \textbf{0.84}  & \textbf{0.85}  & \textbf{0.85} \\ \hline

    \multirow{4}{*}{blood} & 2 & 246.76 &	\textcolor[HTML]{757171}{1.82} & \textcolor[HTML]{757171}{1.84} &	\textcolor[HTML]{757171}{1.89} & \textcolor[HTML]{757171}{1.87} \\
    & 3 & (14.14) & \textcolor[HTML]{757171}{1.35}  & \textcolor[HTML]{757171}{1.28}  & \textcolor[HTML]{757171}{1.25}  & \textcolor[HTML]{757171}{1.26} \\
    & 4 & (21.71) & \textcolor[HTML]{757171}{1.02}  & \textcolor[HTML]{757171}{1.04}  & \textcolor[HTML]{757171}{1.04}  & \textcolor[HTML]{757171}{1.04} \\
    & 5 & (20.97) & \textcolor[HTML]{757171}{1.23}  & \textcolor[HTML]{757171}{1.27}  & \textcolor[HTML]{757171}{1.27}  & \textcolor[HTML]{757171}{1.27} \\ \hline
\end{tabular}}
\end{table}
\normalsize

\section{Discussion}
Here we would like to have a brief discussion on the use of $\milo$ formulations over heuristic or branch and bound formulations for finding optimal binary classification trees as well as a direction for future work related to $\milo$ formulations of binary decision trees.

Observe in Table~\ref{tab:accuracy} there is a drop off in out-of-sample accuracy of branch and bound models for categorical datasets, particularly GOSDT+guesses. In numerical datasets as the tree becomes large DL8.5 encountered memory problems. Our proposed models remain robust enough to handle the varying data structures. With numerical datasets, where the feature space is greatly expanded through the bucketization process of~\citet{lin2020}, one would expect $\milo$ models to perform poorly due to the large search space. However, this is clearly not the case as out-of-sample accuracy remains competitive with branch and bound models and test instances report in-sample optimality gap $<10\%$. In general when optimality gap is $>15\%$ we still see strong out-of-sample performance of our proposed $\milo$ models. This raises an important question for supervised machine learning methods such as decision trees solved used $\milo$ formulations; does optimality need to be achieved? Our models report gap in 27 out of 64 instances, but the branch and bound models outperform in out-of-sample accuracy in only 9 such instances by an average of $4.44\%$. We have not explored the natural question to ask; what percentage gap is \emph{good enough} for sufficient out-of-sample performance.~\citet{aghaei2021} showed a majority of optimality gap is reduced within 1500s for both large and small datasets.

We showed $\baseq$ is TU allowing for the use of LP relaxations to find integral solutions, however our cut-constraints~\eqref{cut13} and~\eqref{cut23}, which are necessary for datapoint path feasiblity, break such integrality guarantees. One future direction we would like to explore is finding path feasibility constraints that keep the matrix TU. We have not explored dual graph or symmetry breaking approaches but would like to in the future.

The Pareto frontiers in Figure~\ref{fig:pareto} cannot be generated using the branch and bound models, as there is no way to control the final number of branching vertices in the tree. These relatively easy to understand plots allow users to have a quick understanding of tree complexity and its relationship to in-sample optimization and out-of-sample test performance. An algorithm that gives control over the final tree topology is important when considering its interpretability and practicality. It is important to note each of the branch and bound benchmark models generates a multivariate decision tree. Complete understanding of a univariate tree is markedly easier than understanding a multivariate tree. Our models do not sacrifice out-of-sample performance for such interpretability.


Regularization can also be applied by adding constraints of the form,
\begin{align*}
    &\sum_{i \in I} s^i_v \geq K_{reg}*p_v & \forall v \in V,
\end{align*}
which ensure each classification vertex contains at least~$K_{reg}$ correctly classified points. Currently we only explore a simple objective of maximizing correctly classified nodes which does not address issues of bias, fairness, or robustness. There is high interest in interpretable models that remain algorithmicly unbiased~\citep{mundru2019, malik2019, fu2020}. To transform our models into such unbiased algorithms we simply change objective function~\eqref{basemax}. 


\section{Conclusion}\label{conclusion}
In this paper we propose four new mixed integer linear optimization formulations that improve upon the recent flow-based formulation $\flowoct$ of~\citet{aghaei2021}. These novel models improve on the linear optimization relaxation of $\flowoct$, shown theoretically and empirically through experimental testing. An improvement in solution time or in-sample optimality gap is observed in 56 out of 64 test instances. Our models are able to outperform a tailored Benders' decomposition of $\flowoct$ in 32 out of 64 test instances. We improve solution times by adding feasibility cuts (fractional and integral) at the root node of the branch and bound tree. Improvement in out-of-sample accuracy over $\flowoct$ and its Benders' decomposition through models $\cutone$ and $\cuttwo$ is observed in 33 out of 64 test instances. Our approach remains competitive against the vast landscape of supervised machine learning models as evidenced by our out-of-sample performance against the most current branch-and-bound methods. Our biobjective approach allows for Pareto frontiers that provide succinct interpretations of tree topology which show our models provide comparatively higher out-of-sample accuracy, even as the tree size arbitrarily increases. Lastly, our models show their robustness as they handle categorical and numerical datasets with both large feature spaces and training sample size.



\bibliographystyle{informs2014} 
\bibliography{DT.bib} 

\newpage
\begin{APPENDICES}
\section{}\label{appendix}
\subsection{Remarks~\ref{flowremark},~\ref{mcf1remark}, and~\ref{mcf2remark} Proofs}\label{appendix:remarks}
\proof{Proof of Remark~\ref{flowremark}.}
For every datapoint $i \in I$, we have
\begin{subequations}
\begin{align}
    \sum_{v \in V} s^i_v &= \sum_{v \in V} s^i_v + 0 \label{implied01}
    \\ 
    &= \sum_{v \in V} s^i_v + \Big(\sum_{v \in V} \Big(z^i(\delta^+(v)) + s^i_v - z^i(\delta^-(v))\Big) - \sum_{v \in V} s^i_v\Big)\label{implied02}
    \\
    &= z^i(\delta^+(1)) + s^i_1 - z^i(\delta^-(1)) + \sum_{v \in V \setminus \{1\}} \Big(z^i(\delta^+(v)) + s^i_v - z^i(\delta^-(v))\Big)\label{implied03}
    \\
    &= z^i(\delta^+(1)) + s^i_1 - z^i(\delta^-(1)) \label{implied04}
    \\
    &= z^i(\delta^+(1)) + s^i_1 \label{implied05}
    \\
    &\le 1.\label{implied06}
\end{align}
\end{subequations}
Here, equality~\eqref{implied02} holds because the summation of all flows is zero in directed graph $G_h$. Equality~\eqref{implied04} holds by constraints~\eqref{sina2}. Equality~\eqref{implied05} holds because the incoming flow to vertex one is zero in directed graph $G_h$. Finally, inequality~\eqref{implied06} holds by constraints~\eqref{sina1}.
\endproof
\vspace{1.5cm}

\proof{Proof of Remark~\ref{mcf1remark}.}
For every datapoint $i \in I$, we have
\begin{subequations}
\begin{align}
    \sum_{v \in V} s^i_v &= \sum_{v \in V} s^i_v + 0\\ 
    &= \sum_{v \in V} s^i_v + \Big(\sum_{v \in V} \Big(z^i(\delta^+(v)) + s^i_v - z^i(\delta^-(v))\Big) - \sum_{v \in V} s^i_v\Big)\label{implied1}\\
    &= z^i(\delta^+(1)) + s^i_1 - z^i(\delta^-(1)) + \sum_{v \in V \setminus \{1\}} \Big(z^i(\delta^+(v)) + s^i_v - z^i(\delta^-(v))\Big)\label{implied2}\\
    &= z^i(\delta^+(1)) + s^i_1 - z^i(\delta^-(1))\label{implied3}\\
    &= q^i_t.\label{implied4}
\end{align}
\end{subequations}
Here, equality~\eqref{implied1} holds because the summation of net flow on all vertices is zero. Equality~\eqref{implied3} holds by constraints~\eqref{mcf12}. Equality~\eqref{implied4} holds by constraints~\eqref{mcf13}. This finishes the proof.
\endproof
\vspace{1.5cm}

\proof{Proof of Remark~\ref{mcf2remark}.}
Let $(\hat{p}, \hat{s}, \hat{b} , \hat{w}, \hat{z}, \hat{q})$ be a point that belongs to $\mcftwo$. We are to show that the point satisfies constraints~\eqref{remark31}. For every vertex $v \in V \setminus \{1\}$ and every datapoint $i \in I$, we have 
\begin{align}
    \hat{z}^{iv}_{a(v),v} - \hat{z}^{iv}_{v,\ell(v)} - \hat{z}^{iv}_{v,r(v)} &= \hat{z}^{iv}(\delta^-(v)) - \hat{z}^{iv}(\delta^+(v)) - \sum_{u \in V} \left(\hat{z}^{iv}(\delta^-(u)) - \hat{z}^{iv}(\delta^+(u))\right) \label{mcf2implied1} \\
    &= - \sum_{u \in V \setminus \{v\}}\left(\hat{z}^{iv}(\delta^-(u)) - \hat{z}^{iv}(\delta^+(u)\right) \label{mcf2implied2} \\
    &= - \left(\hat{z}^{iv}(\delta^-(1)) - \hat{z}^{iv}(\delta^+(1)\right) = \hat{z}_{1,\ell(1)}^{iv} + \hat{z}_{1,r(1)}^{iv} = \hat{s}^i_v. \label{mcf2implied3}
\end{align}
Here, equality~\eqref{mcf2implied1} holds because the summation of the net flows on all vertices of the graph is zero. Then, the last equality~\eqref{mcf2implied3} holds by constraints~\eqref{mcf21}.

Now we are to show that point $(\hat{p}, \hat{s}, \hat{b} , \hat{w}, \hat{z}, \hat{q})$ satisfies constraints~\eqref{remark32}. For every vertex $v \in V \setminus \{1\}$ and every datapoint $i \in I$, we have 
\begin{align}
    \hat{z}^{iv}_{v, \ell(v)} + \hat{z}^{iv}_{v, r(v)} = \hat{z}^{iv}_{a(v),v} - \hat{s}^i_v = \hat{s}^i_v  - \hat{s}^i_v = 0.
\end{align}
Here, the first equality holds by implied equality~\eqref{remark31}. The second equality holds by constraints~\eqref{mcf23.5}. This finishes the proof.
\endproof
\vspace{1.5cm}

\subsection{Lemmata~\ref{cut1lemma} and~\ref{mcf2lemma} Proofs}\label{appendix:lemmas}
\proof{Proof of Lemma~\ref{cut1lemma}.}\label{cut1proof}
($\supseteq$) Let $(\hat{p}, \hat{s}, \hat{b}, \hat{w}, \hat{q}, \hat{z})$ be a point that belongs to polytope $\mcftwoP$. We are to show that point $(\hat{p},\hat{s},\hat{b},\hat{w},\hat{q})$ belongs to polytope $\cutoneP$. It suffices to show that $(\hat{p},\hat{s},\hat{b},\hat{w},\hat{q})$ satisfies constraints~\eqref{cut13}. For every datapoint $i \in I$, every vertex $v \in V \setminus \{1\}$, and every cut vertex $c$ on the path $P_v$, let $R$ be the set of reachable vertices from vertex 1 in tree $G-c$. Then, we have
\begin{subequations}
\begin{align}
    \hat{s}^i_v &= \hat{z}^{iv}_{1, \ell(1)} + \hat{z}^{iv}_{1, r(1)} \label{str21}\\
    &= \hat{z}^{iv}_{1, \ell(1)} + \hat{z}^{iv}_{1, r(1)} + \hat{z}^{iv}(\delta^+(R \setminus \{1\})) - \hat{z}^{iv}(\delta^-(R \setminus \{1\}))  \label{str22}\\
    &\le \hat{z}^{iv}(\delta^+(R)) - 0 \label{str23}\\
    &= \hat{z}^{iv}(\delta^-(c)) \label{str24}\\
    &\le \sum_{u \in V \setminus \{1\}} \hat{z}^{iu}(\delta^-(c))\label{str24.5}\\
    &\le \hat{q}^i_c \label{str25}.
\end{align}
\end{subequations}
Here, equality~\eqref{str21} holds by constraints~\eqref{mcf21}. Equality~\eqref{str22} holds by flow conservation constraints~\eqref{mcf22}. Inequality~\eqref{str23} holds by nonnegativity of $z$ variables. Equality~\eqref{str24} holds because vertex $c$ is a $1,v$-separator. Inequality~\eqref{str24.5} holds by nonnegativity of $z$ variables. Finally, inequality~\eqref{str25} holds by constraints~\eqref{mcf23}.  

($\subseteq$) Let $(\hat{b}, \hat{w}, \hat{s}, \hat{p}, \hat{q})$ be a point that belongs to $\cutoneP$. We are to show that there exists $\bar{z}$ such that $(\hat{b}, \hat{w}, \hat{s}, \hat{p}, \hat{q}, \bar{z})$ belongs to $\mcftwoP$. It suffices to show that the point satisfies constraints~\eqref{mcf21}-\eqref{mcf23.5}. To construct $\bar{z}$, we solve the following max-flow problem for every datapoint $i \in I$ and every vertex $v \in V \setminus \{1\}$.
\vspace{-1ex}
\begin{subequations}
\label{maxflow2}
\begin{align}
\max~&z^{iv}(\delta^+(1)) - z^{iv}(\delta^-(1))  &\label{maxFlow20}\\
\text{s.t.}~& z^{iv}(\delta^+(u)) - z^{iv}(\delta^-(u)) = 0  & \forall u \in V \setminus \{1,v\},~\forall i \in I\label{maxFlow21}\\
& z^{iv}(\delta^-(v)) \le \hat{s}_v^i \label{maxFlow22}\\
& z^{iv}(\delta^-(u)) \le \frac{\hat{q}^i_v}{n - 2}   & \forall u \in V \setminus \{1,v\},~\forall i \in I\label{maxFlow23}\\
&z^{iv}_a \ge 0 & \forall v \in V,~\forall i \in I,~\forall a \in E(G) \label{maxFlow24}.
\end{align}
\end{subequations}
Let $\bar{z}$ be an optimal solution obtained by solving the max-flow problem~\eqref{maxflow2} for every datapoint $i \in I$ and every vertex $u \in V \setminus \{1\}$. It is clear that point $(\hat{b}, \hat{w}, \hat{s}, \hat{p}, \hat{q}, \bar{z})$ satisfies constraints~\eqref{mcf22} because it satisfies constraints~\eqref{maxFlow21}. Furthermore, one can sum up both sides of constraints~\eqref{maxFlow23} on vertices $V \setminus \{1,v\}$. This results in
\begin{align}
    \sum_{u \in V \setminus \{1,v\}} \bar{z}^{iv} (\delta^-(u)) \le \sum_{u \in V \setminus \{1,v\}}  \frac{\hat{q}^i_v}{n - 2} = \hat{q}^i_v. 
\end{align}

Hence, constraints~\eqref{mcf23} are satisfied. Now we show that $(\hat{b}, \hat{w}, \hat{s}, \hat{p}, \hat{q}, \bar{z})$ satisfies constraints~\eqref{mcf21} and~\eqref{mcf23.5}. For every datapoint $i \in I$, every vertex $v \in V \setminus \{1\}$ and every $1,v$-separator $c \in P_v$,
\begin{align*}
    \hat{s}^i_v \le \hat{q}^i_c = \bar{z}^{iv}(\delta^+(1)) - \bar{z}^{iv}(\delta^-(1)) \le \hat{s}^i_v.   
\end{align*}
Here, the first inequality holds by constraints~\eqref{cut13} The first equality holds by the max-flow/min-cut result of~\citet{ford1962}. The last inequality holds because
\begin{align*}
    \bar{z}^{iv}(\delta^+(1)) - \bar{z}^{iv}(\delta^-(1)) &= \bar{z}^{iv}(\delta^+(1)) - \bar{z}^{iv}(\delta^-(1)) - \sum_{u \in V} \left(\bar{z}^{iv}(\delta^+(u)) - \bar{z}^{iv}(\delta^-(u))\right)\\
    &= - \sum_{u \in V \setminus \{1\}} \left(\bar{z}^{iv}(\delta^+(u)) - \bar{z}^{iv}(\delta^-(u))\right)\\
    &= - \left(\bar{z}^{iv}(\delta^+(v)) - \bar{z}^{iv}(\delta^-(v))\right)\\
    &=\bar{z}^{iv}(\delta^-(v)) - \bar{z}^{iv}(\delta^+(v)) \le \bar{z}^{iv}(\delta^-(v)) \le \hat{s}^i_v.
\end{align*}
Here, the first equality holds because $\sum_{u \in V} \left(\bar{z}^{iv}(\delta^+(u)) - \bar{z}^{iv}(\delta^-(u))\right) = 0$ for every datapoint $i \in I$ and every vertex $v \in V \setminus \{1\}$. The third equality holds by constraints~\eqref{maxFlow21}. The last inequality holds by constraints~\eqref{maxFlow22}. This concludes the proof.
\endproof
\vspace{1.5cm}

\proof{Proof of Lemma~\ref{mcf2lemma}.}\label{mcf2proof}
($\subseteq$) Let $(\hat{p}, \hat{s}, \hat{b}, \hat{w}, \hat{q}, \hat{z})$ be a point that belongs to $\mcftwoP$. We are to show that there exist $\bar{z}$ and $\bar{q}$ such that $(\hat{p}, \hat{s}, \hat{b}, \hat{w}, \bar{q}, \bar{z}) \in \mcfoneP$. For every datapoint $i \in I$ and directed edge $(u,v) \in E(G^t_h)$, we define
\begin{align}\label{zDef}
    \bar{z}_{uv}^i := \sum_{p \in V \setminus \{1\}} \hat{z}^{ip}_{uv}. 
\end{align}
For every datapoint $i \in I$ and every vertex $v \in V$, we define $\bar{q}^i_v := \hat{q}^i_v$. For every datapoint $i \in I$ and vertex $t \in V^t$, we define   
\begin{align}\label{uDef}
    \bar{q}^i_t := \sum_{v \in V} \hat{s}^i_v.
\end{align}
We note that $\bar{q}^i_t \le 1$ by constraints~\eqref{mcf25}. It is clear that $(\hat{p}, \hat{s}, \hat{b}, \hat{w}, \bar{z}, \bar{q}) \in \baseqP$. So, it suffices to show that $(\hat{p}, \hat{s}, \hat{b}, \hat{w}, \bar{z}, \bar{q})$ satisfies constraints~\eqref{mcf11}-\eqref{mcf13}. For every datapoint $i \in I$ and every vertex $v \in V \setminus \{1\}$, we show that $(\hat{p}, \hat{s}, \hat{b}, \hat{w}, \bar{z}, \bar{q})$ satisfies constraints~\eqref{mcf11}.
\begin{align*}
    \bar{z}^i_{a(v),v} &= \sum_{u \in V \setminus \{1\}} \hat{z}^{iu}_{a(v),v} \le \hat{q}^i_v = \bar{q}^i_v.  
\end{align*}
Here, the first equality holds by definition~\eqref{zDef}. The first inequality holds by constraints~\eqref{mcf23}. The last equality holds by the definition of $\bar{q}$.


For every datapoint $i \in I$ and for every vertex $v \in V \setminus \{1\}$, we are to show that $(\hat{p}, \hat{s}, \hat{b}, \hat{w}, \bar{z}, \bar{q})$ satisfies constraints~\eqref{mcf12}. 
\begin{align*}
    \bar{z}^i_{a(v),v} - \bar{z}^i_{v, \ell(v)} - \bar{z}^i_{v, r(v)} &= \sum_{u \in V \setminus \{1\}} \hat{z}^{iu}_{a(v),v} - \sum_{u \in V \setminus \{1\}} (\hat{z}^{iu}_{v , \ell(v)} + \hat{z}^{iu}_{v , r(v)}) \\
    &= \sum_{u \in V \setminus \{1,v\}} \hat{z}^{iu}_{a(v),v} - \sum_{u \in V \setminus \{1,v\}} (\hat{z}^{iu}_{v , \ell(v)} + \hat{z}^{iu}_{v , r(v)})\\
    &~~~+ \hat{z}^{iv}_{a(v),v} - ( \hat{z}^{iv}_{v, \ell(v)} + \hat{z}^{iv}_{v, r(v)})\\
    &= 0 + \hat{z}^{iv}_{a(v),v} - ( \hat{z}^{iv}_{v, \ell(v)} + \hat{z}^{iv}_{v, r(v)})\\
    &=0 + \hat{s}^i_v - 0 = \hat{s}^i_v.
\end{align*}
Here, the first equality holds by definition~\eqref{zDef}. The third equality holds by constraints~\eqref{mcf22}. Finally, the last line holds by constraints~\eqref{mcf23.5} and implied constraints~\eqref{remark32}.

For every datapoint $i \in I$, we show that point $(\hat{p}, \hat{s}, \hat{b}, \hat{w}, \bar{z}, \bar{q})$ satisfies constraints~\eqref{mcf13}.
\begin{align*}
    \bar{q}^i_t &= \sum_{v \in V} \hat{s}^i_v \\
    &= \hat{s}^i_1 + \sum_{v \in V \setminus \{1\}} \hat{s}^i_v \\
    &= \hat{s}^i_1 + \sum_{v \in V \setminus \{1\}} \left(\hat{z}^{iv}_{1, \ell(1)} + \hat{z}^{iv}_{1, r(1)} \right) \\
    &= \hat{s}^i_1 + \bar{z}^i_{1, \ell(1)} + \bar{z}^i_{1, r(1)}.
\end{align*}    

Here, the first equality holds by definition~\eqref{uDef}. The third equality holds by constraints~\eqref{mcf21}. Finally, the last equality holds by definition~\eqref{zDef}. 

($\supseteq$) As $\cutoneP = \proj_{X}\mcftwoP$ by Lemma~\ref{cut1lemma}, it suffices to show that $\cutoneP \supseteq \proj_{X}\mcfoneP$. Let $(\hat{p}, \hat{s}, \hat{b}, \hat{w}, \hat{q}, \hat{z})$ be a point that belongs to $\mcfoneP$. We are to show that $(\hat{p}, \hat{s}, \hat{b}, \hat{w}, \hat{q})$ belongs to the $\cutoneP$ polytope. It suffices to show that $(\hat{p}, \hat{s}, \hat{b}, \hat{w}, \hat{q})$ satisfies constraints~\eqref{cut12} and~\eqref{cut13}. For every datapoint $i \in I$, for every vertex $v \in V \setminus \{1\}$, for every cut vertex $c \in V(P_v)$, we have,
\begin{align*}
    \hat{s}^i_v \le \hat{z}^i (\delta^- (v)) \le \hat{z}^i (\delta^- (c)) \le  \hat{q}^i_c.
\end{align*}   
Here, the first and second inequalities hold by flow conservation constraints~\eqref{mcf12}. The last inequality holds by constraints~\eqref{mcf11}.

Now we show that the point satisfies constraints~\eqref{cut13}.
\begin{align*}
    \sum_{v \in V} \hat{s}^i_v = \hat{q}^i_t \le 1.
\end{align*}
Here, the equality holds by the implied constraints of Remark~\ref{mcf2remark}. The inequality holds because $\hat{q}^i_t$ is a binary variable. This completes the proof. 
\endproof

In figure~\ref{fig:polytopeInclusion} are various orientations of a low-dimensional example of how $\cuttwoP$ does not cutoff optimal solutions when the inclusion is not strict. In Figure~\ref{fig:polytopeInclusion} the top face is parallel to the $x-y$ plane, therefore its vertices have the same objective value. This distinction is important in the context of optimal decision trees as more than one tree structure may produce equivalent objective values; ex. two trees with node assignments mirrored across the root node.

\begin{figure}[H]
\centering
\begin{tikzpicture}
\begin{axis}[
  view={15}{35},
  axis lines=center,
  width=7.8cm,height=6.5cm,
  xmin=-5,xmax=10,ymin=-6,ymax=0, zmin=0,zmax=9,
  y label style={at={(axis cs:0,-6,0)}},
  xlabel={$x$},ylabel={$y$},
]
\addplot3 [only marks] coordinates
{
    (0,0,0) 
};

\addplot3 [no marks,black!25] coordinates {(8,0,0) (8,-2,0) (8,-2,2) (8,0,2) (8,0,0)}; 
\addplot3 [no marks,black!25] coordinates {(8,-2,2) (8,0,2) (2,0,8) (2,-2,8) (8,-2,2)}; 
\addplot3 [no marks,black!25] coordinates {(8,-2,0) (8,-2,2) (2,-2,8) (0,-5,8) (0,-5,0) (8,-2,0)}; 
\addplot3 [no marks,black!25] coordinates {(0,-5,8) (0,-5,0) (-3,-3,8) (0,-5,8)}; 
\addplot3 [no marks,black!25] coordinates {(2,0,8) (2,-2,8) (0,-5,8) (-3,-3,8) (-2,0,8) (2,0,8)}; 
\addplot3 [no marks,black!25,densely dashed] coordinates {(-2,0,8) (0,0,0)};

\addplot3 [no marks,blue] coordinates {(0,-5,8) (-3,-3,8) (0,-3,0) (0,-5,8)}; 
\addplot3 [no marks,blue] coordinates {(2,0,8) (2,-2,8) (0,-5,8) (-3,-3,8) (-2,0,8) (2,0,8)}; 
\addplot3 [no marks,blue] coordinates {(2,-2,8) (2,0,8) (5,0,0) (5,-2,0) (2,-2,8)}; 
\addplot3 [no marks,blue] coordinates {(0,0,0) (5,0,0) (5,-2,0) (0,-3,0) (0,0,0)}; 
\addplot3 [no marks,densely dashed] coordinates {(-2,0,8) (0,0,0)};

\node [above right] at (axis cs:0,0,0) {$0$};

\addplot3[patch,patch type=polygon,vertex count=4,line width=0pt,opacity=0,fill opacity=0.15,color=black,faceted color=none]
  table[row sep=crcr, point meta=\thisrow{c}] {%
  x y z c\\
  8 -2 0 0\\
  8 0 0 0\\
  8 0 2 0\\
  8 -2 2 -0\\
  };
\addplot3[patch,patch type=polygon,vertex count=4,line width=0pt,opacity=0,fill opacity=0.15,color=black,faceted color=none]
  table[row sep=crcr, point meta=\thisrow{c}] {%
  x y z c\\
  8 -2 2 0\\
  8 0 2 0\\
  2 0 8 0\\
  2 -2 8 0\\
  };
\addplot3[patch,patch type=polygon,vertex count=5,line width=0pt,opacity=0,fill opacity=0.15,color=black,faceted color=none]
  table[row sep=crcr, point meta=\thisrow{c}] {%
  x y z c\\
  2 -2 8 0\\
  8 -2 2 0\\
  8 -2 0 0\\
  0 -5 0 0\\
  0 -5 8 0\\
  };
\addplot3[patch,patch type=triangle,vertex count=3,line width=0pt,opacity=0,fill opacity=0.15,color=black,faceted color=none]
  table[row sep=crcr, point meta=\thisrow{c}] {%
  x y z c\\
  -3 -3 8 0\\
  0 -5 8 0\\
  0 -5 0 0\\
  };
\addplot3[patch,patch type=polygon,vertex count=5,line width=0pt,opacity=0,fill opacity=0.35,color=blue,faceted color=none]
  table[row sep=crcr, point meta=\thisrow{c}] {%
  x y z c\\
  0 -5 8 0\\
  -3 -3 8 0\\
  -2 0 8 0\\
  2 0 8 0\\
  2 -2 8 0\\
  };
\addplot3[patch,patch type=polygon,vertex count=3,line width=0pt,opacity=0,fill opacity=0.35,color=blue,faceted color=none]
  table[row sep=crcr, point meta=\thisrow{c}] {%
  x y z c\\
  -3 -3 8 0\\
  0 -5 8 0\\
  0 -3 0 0\\
  };
\addplot3[patch,patch type=polygon,vertex count=4,line width=0pt,opacity=0,fill opacity=0.35,color=blue,faceted color=none]
  table[row sep=crcr, point meta=\thisrow{c}] {%
  x y z c\\
  2 -2 8 0\\
  5 -2 0 0\\
  0 -3 0 0\\
  0 -5 8 0\\
  };
\addplot3[patch,patch type=polygon,vertex count=4,line width=0pt,opacity=0,fill opacity=0.35,color=blue,faceted color=none]
  table[row sep=crcr, point meta=\thisrow{c}] {%
  x y z c\\
  2 -2 8 0\\
  2 0 8 0\\
  5 0 0 0\\
  5 -2 0 0\\
  };
\addplot3[patch,patch type=polygon,vertex count=4,line width=0pt,opacity=0,fill opacity=0.35,color=blue,faceted color=none]
  table[row sep=crcr, point meta=\thisrow{c}] {%
  x y z c\\
  0 0 0 0\\
  0 -3 0 0\\
  -3 -3 8 0\\
  -2 0 8 0\\
  };
\addplot3[patch,patch type=polygon,vertex count=4,line width=0pt,opacity=0,fill opacity=0.35,color=blue,faceted color=none]
  table[row sep=crcr, point meta=\thisrow{c}] {%
  x y z c\\
  -2 0 8 0\\
  2 0 8 0\\
  5 0 0 0\\
  0 0 0 0\\
  };
\addplot3[patch,patch type=polygon,vertex count=4,line width=0pt,opacity=0,fill opacity=0.35,color=blue,faceted color=none]
  table[row sep=crcr, point meta=\thisrow{c}] {%
  x y z c\\
  0 0 0 0\\
  0 -3 0 0\\
  5 -2 0 0\\
  5 0 0 0\\
  };
\end{axis}
\end{tikzpicture}
\qquad
\begin{tikzpicture}
\begin{axis}[
  view={15}{0},
  axis lines=center,
  width=7.8cm,height=6.5cm,
  xmin=-5,xmax=10,ymin=-6,ymax=0, zmin=0,zmax=9,
  xlabel={$x$},zlabel={$z$},
]
\addplot3 [only marks] coordinates
{
    (0,0,0) 
};

\addplot3 [no marks,black!25] coordinates {(8,0,0) (8,-2,0) (8,-2,2) (8,0,2) (8,0,0)}; 
\addplot3 [no marks,black!25] coordinates {(8,-2,2) (8,0,2) (2,0,8) (2,-2,8) (8,-2,2)}; 
\addplot3 [no marks,black!25] coordinates {(8,-2,0) (8,-2,2) (2,-2,8) (0,-5,8) (0,-5,0) (8,-2,0)}; 
\addplot3 [no marks,black!25] coordinates {(0,-5,8) (0,-5,0) (-3,-3,8) (0,-5,8)}; 
\addplot3 [no marks,black!25] coordinates {(2,0,8) (2,-2,8) (0,-5,8) (-3,-3,8) (-2,0,8) (2,0,8)}; 
\addplot3 [no marks,black!25,densely dashed] coordinates {(-2,0,8) (0,0,0)};

\addplot3 [no marks,blue] coordinates {(0,-5,8) (-3,-3,8) (0,-3,0) (0,-5,8)}; 
\addplot3 [no marks,blue] coordinates {(2,0,8) (2,-2,8) (0,-5,8) (-3,-3,8) (-2,0,8) (2,0,8)}; 
\addplot3 [no marks,blue] coordinates {(2,-2,8) (2,0,8) (5,0,0) (5,-2,0) (2,-2,8)}; 
\addplot3 [no marks,blue] coordinates {(0,0,0) (5,0,0) (5,-2,0) (0,-3,0) (0,0,0)}; 
\addplot3 [no marks,densely dashed] coordinates {(-2,0,8) (0,0,0)};

\node [above right] at (axis cs:0,0,0) {$0$};

\addplot3[patch,patch type=polygon,vertex count=4,line width=0pt,opacity=0,fill opacity=0.15,color=black,faceted color=none]
  table[row sep=crcr, point meta=\thisrow{c}] {%
  x y z c\\
  8 -2 0 0\\
  8 0 0 0\\
  8 0 2 0\\
  8 -2 2 -0\\
  };
\addplot3[patch,patch type=polygon,vertex count=4,line width=0pt,opacity=0,fill opacity=0.15,color=black,faceted color=none]
  table[row sep=crcr, point meta=\thisrow{c}] {%
  x y z c\\
  8 -2 2 0\\
  8 0 2 0\\
  2 0 8 0\\
  2 -2 8 0\\
  };
\addplot3[patch,patch type=polygon,vertex count=5,line width=0pt,opacity=0,fill opacity=0.15,color=black,faceted color=none]
  table[row sep=crcr, point meta=\thisrow{c}] {%
  x y z c\\
  2 -2 8 0\\
  8 -2 2 0\\
  8 -2 0 0\\
  0 -5 0 0\\
  0 -5 8 0\\
  };
\addplot3[patch,patch type=triangle,vertex count=3,line width=0pt,opacity=0,fill opacity=0.15,color=black,faceted color=none]
  table[row sep=crcr, point meta=\thisrow{c}] {%
  x y z c\\
  -3 -3 8 0\\
  0 -5 8 0\\
  0 -5 0 0\\
  };
\addplot3[patch,patch type=polygon,vertex count=5,line width=0pt,opacity=0,fill opacity=0.35,color=blue,faceted color=none]
  table[row sep=crcr, point meta=\thisrow{c}] {%
  x y z c\\
  0 -5 8 0\\
  -3 -3 8 0\\
  -2 0 8 0\\
  2 0 8 0\\
  2 -2 8 0\\
  };
\addplot3[patch,patch type=polygon,vertex count=3,line width=0pt,opacity=0,fill opacity=0.35,color=blue,faceted color=none]
  table[row sep=crcr, point meta=\thisrow{c}] {%
  x y z c\\
  -3 -3 8 0\\
  0 -5 8 0\\
  0 -3 0 0\\
  };
\addplot3[patch,patch type=polygon,vertex count=4,line width=0pt,opacity=0,fill opacity=0.35,color=blue,faceted color=none]
  table[row sep=crcr, point meta=\thisrow{c}] {%
  x y z c\\
  2 -2 8 0\\
  5 -2 0 0\\
  0 -3 0 0\\
  0 -5 8 0\\
  };
\addplot3[patch,patch type=polygon,vertex count=4,line width=0pt,opacity=0,fill opacity=0.35,color=blue,faceted color=none]
  table[row sep=crcr, point meta=\thisrow{c}] {%
  x y z c\\
  2 -2 8 0\\
  2 0 8 0\\
  5 0 0 0\\
  5 -2 0 0\\
  };
\addplot3[patch,patch type=polygon,vertex count=4,line width=0pt,opacity=0,fill opacity=0.35,color=blue,faceted color=none]
  table[row sep=crcr, point meta=\thisrow{c}] {%
  x y z c\\
  0 0 0 0\\
  0 -3 0 0\\
  -3 -3 8 0\\
  -2 0 8 0\\
  };
\addplot3[patch,patch type=polygon,vertex count=4,line width=0pt,opacity=0,fill opacity=0.35,color=blue,faceted color=none]
  table[row sep=crcr, point meta=\thisrow{c}] {%
  x y z c\\
  -2 0 8 0\\
  2 0 8 0\\
  5 0 0 0\\
  0 0 0 0\\
  };
\addplot3[patch,patch type=polygon,vertex count=4,line width=0pt,opacity=0,fill opacity=0.35,color=blue,faceted color=none]
  table[row sep=crcr, point meta=\thisrow{c}] {%
  x y z c\\
  0 0 0 0\\
  0 -3 0 0\\
  5 -2 0 0\\
  5 0 0 0\\
  };
\end{axis}
\end{tikzpicture}
\qquad
\\
\begin{tikzpicture}
\begin{axis}[
  view={75}{45},
  axis lines=center,
  width=7cm,height=5.833cm,
  xmin=-5,xmax=10,ymin=-6,ymax=0, zmin=0,zmax=9,
  y label style={at={(axis cs:0,-6,0)}},
  xlabel={$x$},ylabel={$y$},
]
\addplot3 [only marks] coordinates
{
    (0,0,0) 
};

\addplot3 [no marks,black!25] coordinates {(8,0,0) (8,-2,0) (8,-2,2) (8,0,2) (8,0,0)}; 
\addplot3 [no marks,black!25] coordinates {(8,-2,2) (8,0,2) (2,0,8) (2,-2,8) (8,-2,2)}; 
\addplot3 [no marks,black!25] coordinates {(8,-2,0) (8,-2,2) (2,-2,8) (0,-5,8) (0,-5,0) (8,-2,0)}; 
\addplot3 [no marks,black!25] coordinates {(0,-5,8) (0,-5,0) (-3,-3,8) (0,-5,8)}; 
\addplot3 [no marks,black!25] coordinates {(2,0,8) (2,-2,8) (0,-5,8) (-3,-3,8) (-2,0,8) (2,0,8)}; 
\addplot3 [no marks,black!25,densely dashed] coordinates {(-2,0,8) (0,0,0)};

\addplot3 [no marks,blue] coordinates {(0,-5,8) (-3,-3,8) (0,-3,0) (0,-5,8)}; 
\addplot3 [no marks,blue] coordinates {(2,0,8) (2,-2,8) (0,-5,8) (-3,-3,8) (-2,0,8) (2,0,8)}; 
\addplot3 [no marks,blue] coordinates {(2,-2,8) (2,0,8) (5,0,0) (5,-2,0) (2,-2,8)}; 
\addplot3 [no marks,blue] coordinates {(0,0,0) (5,0,0) (5,-2,0) (0,-3,0) (0,0,0)}; 
\addplot3 [no marks,densely dashed] coordinates {(-2,0,8) (0,0,0)};

\node [above right] at (axis cs:0,0,0) {$0$};

\addplot3[patch,patch type=polygon,vertex count=4,line width=0pt,opacity=0,fill opacity=0.15,color=black,faceted color=none]
  table[row sep=crcr, point meta=\thisrow{c}] {%
  x y z c\\
  8 -2 0 0\\
  8 0 0 0\\
  8 0 2 0\\
  8 -2 2 -0\\
  };
\addplot3[patch,patch type=polygon,vertex count=4,line width=0pt,opacity=0,fill opacity=0.15,color=black,faceted color=none]
  table[row sep=crcr, point meta=\thisrow{c}] {%
  x y z c\\
  8 -2 2 0\\
  8 0 2 0\\
  2 0 8 0\\
  2 -2 8 0\\
  };
\addplot3[patch,patch type=polygon,vertex count=5,line width=0pt,opacity=0,fill opacity=0.15,color=black,faceted color=none]
  table[row sep=crcr, point meta=\thisrow{c}] {%
  x y z c\\
  2 -2 8 0\\
  8 -2 2 0\\
  8 -2 0 0\\
  0 -5 0 0\\
  0 -5 8 0\\
  };
\addplot3[patch,patch type=triangle,vertex count=3,line width=0pt,opacity=0,fill opacity=0.15,color=black,faceted color=none]
  table[row sep=crcr, point meta=\thisrow{c}] {%
  x y z c\\
  -3 -3 8 0\\
  0 -5 8 0\\
  0 -5 0 0\\
  };
\addplot3[patch,patch type=polygon,vertex count=5,line width=0pt,opacity=0,fill opacity=0.35,color=blue,faceted color=none]
  table[row sep=crcr, point meta=\thisrow{c}] {%
  x y z c\\
  0 -5 8 0\\
  -3 -3 8 0\\
  -2 0 8 0\\
  2 0 8 0\\
  2 -2 8 0\\
  };
\addplot3[patch,patch type=polygon,vertex count=3,line width=0pt,opacity=0,fill opacity=0.35,color=blue,faceted color=none]
  table[row sep=crcr, point meta=\thisrow{c}] {%
  x y z c\\
  -3 -3 8 0\\
  0 -5 8 0\\
  0 -3 0 0\\
  };
\addplot3[patch,patch type=polygon,vertex count=4,line width=0pt,opacity=0,fill opacity=0.35,color=blue,faceted color=none]
  table[row sep=crcr, point meta=\thisrow{c}] {%
  x y z c\\
  2 -2 8 0\\
  5 -2 0 0\\
  0 -3 0 0\\
  0 -5 8 0\\
  };
\addplot3[patch,patch type=polygon,vertex count=4,line width=0pt,opacity=0,fill opacity=0.35,color=blue,faceted color=none]
  table[row sep=crcr, point meta=\thisrow{c}] {%
  x y z c\\
  2 -2 8 0\\
  2 0 8 0\\
  5 0 0 0\\
  5 -2 0 0\\
  };
\addplot3[patch,patch type=polygon,vertex count=4,line width=0pt,opacity=0,fill opacity=0.35,color=blue,faceted color=none]
  table[row sep=crcr, point meta=\thisrow{c}] {%
  x y z c\\
  0 0 0 0\\
  0 -3 0 0\\
  -3 -3 8 0\\
  -2 0 8 0\\
  };
\addplot3[patch,patch type=polygon,vertex count=4,line width=0pt,opacity=0,fill opacity=0.35,color=blue,faceted color=none]
  table[row sep=crcr, point meta=\thisrow{c}] {%
  x y z c\\
  -2 0 8 0\\
  2 0 8 0\\
  5 0 0 0\\
  0 0 0 0\\
  };
\addplot3[patch,patch type=polygon,vertex count=4,line width=0pt,opacity=0,fill opacity=0.35,color=blue,faceted color=none]
  table[row sep=crcr, point meta=\thisrow{c}] {%
  x y z c\\
  0 0 0 0\\
  0 -3 0 0\\
  5 -2 0 0\\
  5 0 0 0\\
  };
\end{axis}
\end{tikzpicture}
\qquad
\begin{tikzpicture}
\begin{axis}[
  view={0}{15},
  axis lines=center,
  width=7cm,height=5.833cm,
  xmin=-5,xmax=10,ymin=-6,ymax=0, zmin=0,zmax=9,
  xlabel={$x$},zlabel={$z$},
]
\addplot3 [only marks] coordinates
{
    (0,0,0) 
};

\addplot3 [no marks,black!25] coordinates {(8,0,0) (8,-2,0) (8,-2,2) (8,0,2) (8,0,0)}; 
\addplot3 [no marks,black!25] coordinates {(8,-2,2) (8,0,2) (2,0,8) (2,-2,8) (8,-2,2)}; 
\addplot3 [no marks,black!25] coordinates {(8,-2,0) (8,-2,2) (2,-2,8) (0,-5,8) (0,-5,0) (8,-2,0)}; 
\addplot3 [no marks,black!25] coordinates {(0,-5,8) (0,-5,0) (-3,-3,8) (0,-5,8)}; 
\addplot3 [no marks,black!25] coordinates {(2,0,8) (2,-2,8) (0,-5,8) (-3,-3,8) (-2,0,8) (2,0,8)}; 
\addplot3 [no marks,black!25,densely dashed] coordinates {(-2,0,8) (0,0,0)};

\addplot3 [no marks,blue] coordinates {(0,-5,8) (-3,-3,8) (0,-3,0) (0,-5,8)}; 
\addplot3 [no marks,blue] coordinates {(2,0,8) (2,-2,8) (0,-5,8) (-3,-3,8) (-2,0,8) (2,0,8)}; 
\addplot3 [no marks,blue] coordinates {(2,-2,8) (2,0,8) (5,0,0) (5,-2,0) (2,-2,8)}; 
\addplot3 [no marks,blue] coordinates {(0,0,0) (5,0,0) (5,-2,0) (0,-3,0) (0,0,0)}; 
\addplot3 [no marks,densely dashed] coordinates {(-2,0,8) (0,0,0)};

\node [above right] at (axis cs:0,0,0) {$0$};

\addplot3[patch,patch type=polygon,vertex count=4,line width=0pt,opacity=0,fill opacity=0.15,color=black,faceted color=none]
  table[row sep=crcr, point meta=\thisrow{c}] {%
  x y z c\\
  8 -2 0 0\\
  8 0 0 0\\
  8 0 2 0\\
  8 -2 2 -0\\
  };
\addplot3[patch,patch type=polygon,vertex count=4,line width=0pt,opacity=0,fill opacity=0.15,color=black,faceted color=none]
  table[row sep=crcr, point meta=\thisrow{c}] {%
  x y z c\\
  8 -2 2 0\\
  8 0 2 0\\
  2 0 8 0\\
  2 -2 8 0\\
  };
\addplot3[patch,patch type=polygon,vertex count=5,line width=0pt,opacity=0,fill opacity=0.15,color=black,faceted color=none]
  table[row sep=crcr, point meta=\thisrow{c}] {%
  x y z c\\
  2 -2 8 0\\
  8 -2 2 0\\
  8 -2 0 0\\
  0 -5 0 0\\
  0 -5 8 0\\
  };
\addplot3[patch,patch type=triangle,vertex count=3,line width=0pt,opacity=0,fill opacity=0.15,color=black,faceted color=none]
  table[row sep=crcr, point meta=\thisrow{c}] {%
  x y z c\\
  -3 -3 8 0\\
  0 -5 8 0\\
  0 -5 0 0\\
  };
\addplot3[patch,patch type=polygon,vertex count=5,line width=0pt,opacity=0,fill opacity=0.35,color=blue,faceted color=none]
  table[row sep=crcr, point meta=\thisrow{c}] {%
  x y z c\\
  0 -5 8 0\\
  -3 -3 8 0\\
  -2 0 8 0\\
  2 0 8 0\\
  2 -2 8 0\\
  };
\addplot3[patch,patch type=polygon,vertex count=3,line width=0pt,opacity=0,fill opacity=0.35,color=blue,faceted color=none]
  table[row sep=crcr, point meta=\thisrow{c}] {%
  x y z c\\
  -3 -3 8 0\\
  0 -5 8 0\\
  0 -3 0 0\\
  };
\addplot3[patch,patch type=polygon,vertex count=4,line width=0pt,opacity=0,fill opacity=0.35,color=blue,faceted color=none]
  table[row sep=crcr, point meta=\thisrow{c}] {%
  x y z c\\
  2 -2 8 0\\
  5 -2 0 0\\
  0 -3 0 0\\
  0 -5 8 0\\
  };
\addplot3[patch,patch type=polygon,vertex count=4,line width=0pt,opacity=0,fill opacity=0.35,color=blue,faceted color=none]
  table[row sep=crcr, point meta=\thisrow{c}] {%
  x y z c\\
  2 -2 8 0\\
  2 0 8 0\\
  5 0 0 0\\
  5 -2 0 0\\
  };
\addplot3[patch,patch type=polygon,vertex count=4,line width=0pt,opacity=0,fill opacity=0.35,color=blue,faceted color=none]
  table[row sep=crcr, point meta=\thisrow{c}] {%
  x y z c\\
  0 0 0 0\\
  0 -3 0 0\\
  -3 -3 8 0\\
  -2 0 8 0\\
  };
\addplot3[patch,patch type=polygon,vertex count=4,line width=0pt,opacity=0,fill opacity=0.35,color=blue,faceted color=none]
  table[row sep=crcr, point meta=\thisrow{c}] {%
  x y z c\\
  -2 0 8 0\\
  2 0 8 0\\
  5 0 0 0\\
  0 0 0 0\\
  };
\addplot3[patch,patch type=polygon,vertex count=4,line width=0pt,opacity=0,fill opacity=0.35,color=blue,faceted color=none]
  table[row sep=crcr, point meta=\thisrow{c}] {%
  x y z c\\
  0 0 0 0\\
  0 -3 0 0\\
  5 -2 0 0\\
  5 0 0 0\\
  };
\end{axis}
\end{tikzpicture}

\caption{The polytope in \textcolor{blue}{blue} is contained in the polytope in grey, analogous to that of $\cuttwoP\subseteq\cutoneP$, even though the top face is shared and faces overlap on the $x-y$, $x-z$, and $y-z$ planes.\label{fig:polytopeInclusion}}
\end{figure}
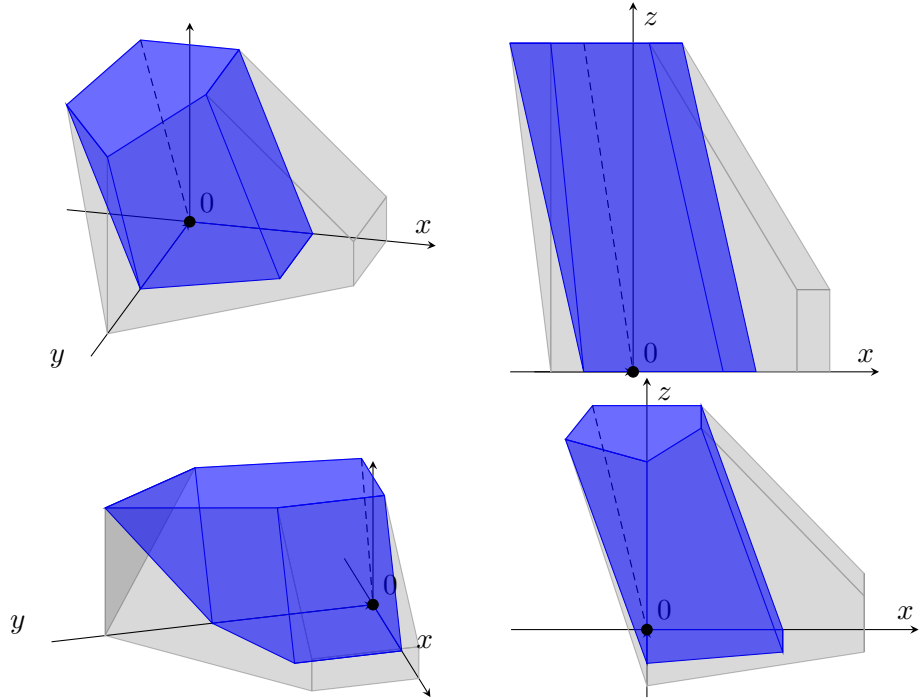
\vspace{1.5cm}

\subsection{Theorems~\ref{flowcorrect} and~\ref{cutcorrect} Proofs}\label{appendix:theorems}
\proof{Proof of Theorem~\ref{flowcorrect}.}\label{cutproof}
Let $p^* \in \flowoctP$ be defined by binary variables $b, w, p, s, q$. We are to show $p^*$ represents an optimal binary classification tree. Variables $b, w, p$ are defined on a vertex set $B \subseteq V$ and $L \subseteq V$ where $|V| = 2^{h+1}-1$ and $h \geq 1$ is integral and represents the maximal depth of a vertex in an unassigned binary decision tree. Further $B \cap L = \emptyset$. On each vertex $v \in V$ the left and right children are defined as $\ell(v), r(v)$ and the set of non-root vertices on the unique $1,v$-path as $P_{1,v}$. Decision variables $b$ and $w$ are defined on binary encoded feature set $F$ and discrete class set $K$, respectively, of given training dataset $I$.
Constraints~\eqref{base1}-\eqref{base3} impose a valid tree structure through the following. For each $p_v=1$, then $\sum_{k \in K} w_{vk}=1$ by~\eqref{base1}. Thus exactly one $k \in K$ was chosen for vertex $v$ (exactly one class was assigned to vertex v). Observe,
\begin{align*}
\sum_{f \in F} b_{vf} = \sum_{u \in V(P_{1,v} \setminus \{v\})} p_u=0.
\end{align*}
This implies, at vertex $v$, it is not assigned any branching features and $p_u$ = 0 for all of the ancestors $u$ of $v$. Given that $p_u = 0$ for each ancestor $u$ of $v$, then $\sum_{f \in F} b_{uf} = 1$ by~\eqref{base2}. This implies each ancestor $u$ of $v$ is assigned exactly one branching feature. Then again by~\eqref{base1}, $\sum_{k \in K} w_{uk} = 0$. This implies each ancestor $u$ of $v$ is not assigned any class. Then if we look at the children of $v$ where $p_v=1$, by~\eqref{base2} we get,
\begin{align*}
    & p_c = \sum_{f \in F} b_{cf} = \sum_{k \in K} w_{ck} = 0 &~\forall c \in \child(v).
\end{align*}
This implies all children $c$ of $v$ are not assigned a class or branching feature, and therefore pruned. If $v \in L$ then $\sum_{f \in F} b_{vf} = 0$ by~\eqref{base3}. This implies no leaf vertex is assigned to a branching feature and also assigned exactly one class. The above statements have shown that a prediction vertex $v \in V$ is assigned exactly one class, all of its ancestors were assigned a single branching feature, all of its children were pruned and that no leaf vertex can be assigned a branching feature.
For each $b_{vf} = 1$, at $v$, 
\begin{align*}
    \sum_{g \in F \setminus \{f\}} b_{vg} = \sum_{u \in V(P_{1,v}) \setminus \{v\}} p_u = p_v = 0,
\end{align*}
by~\eqref{base2}. Then for each ancestor $u$ of $v$, $\sum_{f \in F} b_{uf} = 1$ by~\eqref{base2}. Thus, if vertex $v$ is assigned branching feature $f \in F$, then it is not pruned and each of its ancestors was assigned exactly one branching feature. Then if we look at the children of $c$ of $v$ then either $\sum_{f \in F} b_{cf} = 1$ or $p_c = 1$ by~\eqref{base2}, implying the children of branching vertex $v$ are either branching or assigned a class. For each $w_{vk} = 1$, $p_v = 1$ by~\eqref{base1} and follows the implications previously explicated. The following statements have shown $\flowoctP$ admits a valid assigned binary tree structure. 

Lastly we show $\flowoctP$ admits valid paths through the tree. For each $s^i_v = 1$, by~\eqref{base4} $w_{vk} = 1$ where $k = y^i$. This implies a datapoint classified at vertex $v$ is correct only if the vertex $v$ is assigned the datapoint's given class $k$. Then by constraints~\eqref{sina1} and~\eqref{sina2} the datapoint flows $z^i_a = 1$ for $a \in E(G[V(P_{1,v})])$ along the $1,v$-path are connected and all other flows $z^i_{uv} \in E(G \setminus V(P_{1,v})) = 0$. Thus a correctly classified datapoint found a successful $s,t$ flow in the tree. Then by~\eqref{sina3} the connected flows $z^i_a$ along arcs $a \in E(G[V(P_{1,v})])$ are branched such that the flow $z^i_{v,\ell(v)}$ ($z^i_{v,r(v)}$) along the arc connecting vertex $v$ to its left (right) child, $\ell(v)$ ($r(v)$), meant vertex $v$ was assigned some branching feature $f \in F$ such that $x^i_f = 0~(x^i_f = 1)$. In short, this implies connected flows $z^i_a$ of a datapoint $i \in I$ are such that the flow's left, right branching corresponds to features assigned to branching vertices of the tree and the datapoint's feature set values. Point $p^*$ is an optimal classification tree because it satisfies~\eqref{basemax}, and the maximal number of training datapoints were correctly classified by the tree. This complete the proof.
\endproof
\vspace{1.5cm}

\proof{Proof of Theorem~\ref{cutcorrect}.}\label{flowproof}
Let $p^*$ be a point that represents an optimal binary classification tree trained by a binary encoded, discrete class dataset, $I \coloneqq \{x^i,y^i\}_{i \in |N|}$. We must show $p^* = (p,b,w,s,q)$ is in $\cuttwoP$. Given that $p^*$ is a binary tree starting at the root find the deepest classification vertex at depth $h$. We will assume the tree is of depth $h$. Enumerate vertices starting from the root of the tree, given number 1, until $2^{h+1}-1$ ensuring the left, right children of vertex $v$ are labeled $2v,~2v+1$, respectively. Further, if a vertex is the child of a classification vertex, assume it still exists and is pruned; assume the same for children of pruned vertices. Denote leaf set $L$ as vertices whose label is in $[2^h,~2^{h+1}-1]$, all other vertices as branching set $B \coloneqq [1,~2^h-1]$, and $V = B \cup L$.

For each classification vertex labeled $v$ with class $k \in K$ denote binary variables $w_{vk} = p_v = 1$ and binary variables $w_{vj} = b_{vf} = 0$ for every feature $f \in F$ and $j \in K \setminus \{k\}$, where $F$ and $K$ are the set of training features and classes, respectively. Similarly for every vertex labeled $v$ with branching feature $f$ we denote binary variable $b_{vf} = 1$ and $w_{vk} = b_{vg} = p_v = 0$ for every class $k \in K$ and $g \in F \setminus \{f\}$. For every pruned vertex labeled $v$ we denote binary variable $p_v = 0$ and $w_{vk} = b_{vf} = 0$ for every feature $f \in F$ and $k \in K$.
For every datapoint $i$ in training set $I$ starting at the root with label $1$ we denote binary variable $q^i_1 = 1$. We then branch left right using the $x^i_f$ binary value given to the feature of the branching vertex $v$. From there we denote binary variables $q^i_v = 1$ for every vertex $v$ visited by $i$. All other vertices $v' \in V \setminus \{v\}$ not visited have binary variables $q^i_{v'} = 0$. When datapoint $i$ reaches a classification vertex $u$, if it is correctly assigned we denote binary variable $s^i_u = 1$, else $s^i_u = 0$. We also denote binary variables $s^i_{u'} = 0$ for every $u' \in V \setminus \{u\}$.

For every vertex $v \in V$ with $w_{vk} = 1$ for some $k \in K$, since $w_{vj} = b_{vf} = 0$ for every class $j \in K \setminus \{k\}$ and feature $f \in F$ (and $p_v = 1$), vertex $v$ satisfies constraints~\eqref{base1} and~\eqref{base2}. We assume $p^*$ is a classification tree, thus for every vertex $v \in L$, $v$ must be assigned some class $k \in K$ or pruned. If some $v \in L$ is assigned a class $k$, then $w_{vk} = 1$ and we have the above. Thus, all such $v \in L$ satisfy constraints~\eqref{base3}. For every vertex $v \in V$ with $b_{vf} = 1$ for some $f \in F$, since
\begin{align*}
    &b_{vg} = w_{vk} = p_v = 0 &~\forall g \in F \setminus \{f\},~\forall k \in K,
\end{align*}
every such $v$ satisfies constraints~\eqref{base1} and~\eqref{base2}. There are no such $v \in L$ with $b_{vf} = 1$ for any $f \in F$ since $p^*$ is a classification tree and thus constraints~\eqref{base3} are also satisfied. For each $v$ with $p_v = 1$, then $w_{vk} = 1$ for some $k \in K$ and follows the above. For each $p_v = 0$ then $b_{vf} = 1$ for some $f \in F$ or $w_{vk} = 0$ for all $k \in K$ each following the above. For each datapoint $i \in I$ there will exist a connected $1,v$-path to some terminal vertex $v$, denoted $P_{1,v}$, using the feature set of $x^i$. Let $q^i_c = 1$ for every $c \in P_{1,v}$. Further, since binary variables $b,~w,~p$ satisfy constraints~\eqref{base1}-\eqref{base3} and branching are done using the values of $x^i$, then,
\begin{align*}
q^i_{\ell(c)} \leq \sum_{f \in F:x^i_f=0} b_{cf} \leq \sum_{f \in F} b_{cf} \leq 1,\\
q^i_{r(c)} \leq \sum_{f \in F:x^i_f=1} b_{cf} \leq \sum_{f \in F} b_{cf} \leq 1,
\end{align*}
for each vertex $c \in P_{1,v}$ and for every datapoint $i \in I$. Thus, constraints~\eqref{mcf14} are satisfied for every $i \in I$ and $v \in V$. Lastly for every $i \in I$ and $v \in V$ where $s^i_v=1$, it follows $s^i_v=w_{vk=y^i}=1$, thus constraints~\eqref{base4} are satisfied by construction.

Here we have shown, $p^* \in \baseqP$. Next we show $p*$ satisfies our $\cuttwo$ connectivity constraints. Terminal vertex $v$ was reached through a connected path for each datapoint $i \in I$. Thus, $s^i_v \leq q^i_c$ for every $c \in P_v$ because $s^i_v = q^i_c = 1$ at $v$ and every $c \in P_v$. From our labeling of binary variables $s^i_n = 0$ for every $n \in V \setminus \{v\}$ which implies
\begin{align*}
    s^i_v + \sum_{u \in \child(v)} s^i_u \leq s^i_v + \sum_{u \in V \setminus \{v\}} & s^i_u = \sum_{v \in V} s^i_v = 1
\end{align*}
thereby satisfying constraints~\eqref{cut22} and~\eqref{cut23}. This completes the proof.
\endproof

\subsection{Examples of $\baseq$ Equitable Bicoloring}\label{appendix:bicoloring}
\renewcommand{\arraystretch}{1.5}
\begin{table}[H]
\caption{Examples for $\baseq$ column coloring rules.}\label{tab:bicoloringexamples}
\begin{tabular}{|p{0.13\linewidth}|p{0.77\linewidth}|}
    \hline
     Cols. of D.V. & Labeling Rules\\ \hline
     $p$ & A depth 2 tree has 7 vertices. The columns associated with decision variables $p$ are labeled $\{p_1,\dots,p_7\} \stackrel{L}{=} \{r,b,b,r,r,r,r\}$.\\ \hline
     $w$ & Let $|K|=3$. Columns of of $w$ associated vertices $v={2,3}$ are labeled $\{w_{v1}, w_{v2}, w_{v3}\}\stackrel{L}{=}\{b,r,b\}$. Columns associated with vertices $u\in\{4,5,6,7\}$ are labeled $\{w_{u1}, w_{u2}, w_{u3}\}\stackrel{L}{=}\{r,b,r\}$. \\\hline
     $s$ & Consider datapoint classes $y=\{1,1,2,3\}$. Since $|K|=3$, we know that $\{w_{11}, w_{12}, w_{13}\}$ are labeled $\{r,b,r\}.$ Thus the labels of $\{s^1_1,s^2_1,s^3_1,s^4_1\}\stackrel{L}{=}\{r,r,b,r\}$. The set $\{s^1_2, s^2_2, s^3_2, s^4_2\}$ is labeled $\{b,b,r,b\}$ as $\{w_{21}, w_{22}, w_{23}\}\stackrel{L}{=}\{b,r,b\}$. Now let $\bar{y}=\{1,2,2,3\}$, we then get column labels of $\{\bar{s}^1_1,\bar{s}^2_1,\bar{s}^3_1,\bar{s}^4_1\}\stackrel{L}{=}\{r,b,b,r\}$ and $\{\bar{s}^1_2,\bar{s}^2_2,\bar{s}^3_2,\bar{s}^4_2\}\stackrel{L}{=}\{b,r,r,b\}$. \\\hline
     $b$ & Let $|F|=58,~H=\{a,~b_1,\dots,b_{13},~c_1,c_2,c_3,~d_1,d_2,d_3,d_4\}$. Let $|h|_F$ be the number of encoded features of $h\in H$ observed in $F$ and $|a|_F=7$, $|b_i|_F=2~\forall i\in\{1,\dots,13\}$, $|c_j|_F=3~\forall j\in\{1,2,3\}$ and $|d_k|_F=4~\forall k\in\{1,2,3,4\}$. Using step 1 we pair the $d_k$'s to yield labels of $blue$ for the encoded features of $d_1$ and $d_3$, and $red$ for the encoded features of $d_2$ and $d_4$. We also pair 12 of the $b_i$'s to label $blue$ for the encoded feature of $b_i$ where $i$ is even, and $red$ for encoded features of $b_i$ where $i$ is odd, except $b_{13}$. Lastly we can pair $c_1$ and $c_2$, labeling, arbitrarily, one $red$ and the other $blue$. Left in $H$ unlabeled is $\{a, b_{13}, c_3\} \stackrel{|\cdot|_F}{=} \{7,2,3\}$ for the number of encoded features unlabeled. However, it is clear there is no biparition with a difference of at most 1 in the number of encoded features. Instead, we should not label the pair $c_1, c_2$ to yield an unlabeled set of $\{a,b_{13},c_1,c_2,c_3\} \stackrel{|\cdot|_F}{=} \{7,2,3,3,3\}$. It is now clear the favorable bipartition is $\{a,b_{13}\}$ and $\{c_1,c_2,c_3\}$ where the encoded features of $\{a,b_{13}\}$ are labeled $blue$, and the encoded features of $\{c_1,c_2,c_3\}$ are labeled $red$. \\\hline
     $q$ & Consider a dataset where $H=\{h_i\}_{i=1}^9,~|h_i|_F=7~\forall k$. Following the coloring procedure for $B$ we get the encoded features of $h_{\{1,3,5,7\}}$ are labeled $red$ and the encoded features of $h_{\{2,4,6,8\}}$ are labeled $blue$. Lastly $F_9 \stackrel{L}{=} \{b,r,b,r,b,r,b\}$. For any datapoint, $q^i_{l(v)} \stackrel{L}{=}b$. The labels of $q^i_{r(v)}$ will be dependent on where the 1 exists in $F_9$. If the -1 exists in a column labeled $blue$, then the bicoloring value for columns of $b$ is -1, favoring $blue$ and $q^i_{r(v)}\stackrel{L}{=}b$ for an equitable bicoloring; similarly for a -1 in column labeled $red$.\\\hline
\end{tabular}
\end{table}
\renewcommand{\arraystretch}{1}

\end{APPENDICES}

\end{document}